\documentclass{article}

\usepackage{microtype}
\usepackage{graphicx}
\usepackage{subcaption}
\usepackage{booktabs} 
\usepackage{colortbl} 
\usepackage{float}
\usepackage{multirow}
\usepackage{bm}

\usepackage{hyperref}
\usepackage{hyphenat}
\usepackage{pifont}
\usepackage{xcolor}
\usepackage{tcolorbox}
\definecolor{best}{RGB}{200,215,235}
\definecolor{second}{RGB}{215,220,225}
\newcommand{\highlightnote}{\colorbox{best}{\textbf{Best}} and \colorbox{second}{\underline{second best}} results are highlighted.}

\newcommand{\cmark}{\textcolor{teal}{\ding{51}}}
\newcommand{\xmark}{\textcolor{red}{\ding{55}}}

\newcommand{\ours}{AHAP}

\usepackage[preprint]{icml2026}

\usepackage{amsmath}
\usepackage{amssymb}
\usepackage{mathtools}
\usepackage{amsthm}

\usepackage[capitalize,noabbrev]{cleveref}

\theoremstyle{plain}

\theoremstyle{definition}

\theoremstyle{remark}

\usepackage[textsize=tiny]{todonotes}

\icmltitlerunning{Submission and Formatting Instructions for ICML 2026}

\begin{document}

\twocolumn[
  \icmltitle{AHAP:~Reconstructing Arbitrary Humans from Arbitrary Perspectives \\ with Geometric Priors}

  \icmlsetsymbol{equal}{*}
  \icmlsetsymbol{corresponding}{\textsuperscript{\textdagger}}

  \begin{icmlauthorlist}
    \icmlauthor{Xiaozhen Qiao}{TeleAI,USTC,equal}
    \icmlauthor{Wenjia Wang}{HKU,equal}
    \icmlauthor{Zhiyuan Zhao}{TeleAI}
    \icmlauthor{Jiacheng Sun}{HW}
    \icmlauthor{Ping Luo}{HKU} \\
    \icmlauthor{Hongyuan Zhang}{TeleAI,HKU,corresponding}
    \icmlauthor{Xuelong Li}{TeleAI,corresponding}
  \end{icmlauthorlist}

  \icmlaffiliation{TeleAI}{Institute of Artificial Intelligence (TeleAI), China Telecom, P. R. China.}
  \icmlaffiliation{USTC}{University of Science and Technology of China.}
  \icmlaffiliation{HKU}{University of Hong Kong.}
  \icmlaffiliation{HW}{Huawei Technologies Co., Ltd.}
  \icmlcorrespondingauthor{Hongyuan Zhang}{hyzhang98@gmail.com}
  \icmlcorrespondingauthor{Xuelong Li}{xuelong\_li@ieee.org}

  \icmlkeywords{Machine Learning, ICML}

  \vskip 0.1in
  \begin{center}
    \textbf{Project Page:} \href{https://wenjiawang0312.github.io/projects/ahap/}{\textcolor{teal}{\textbf{AHAP}}} \quad \small{\textsuperscript{*} Equal Contribution \quad \textsuperscript{\textdagger} Corresponding Authors}
  \end{center}
  \vskip 0.2in
]

\printAffiliationsAndNotice{}  

\begin{abstract}
Reconstructing 3D humans from images captured at multiple perspectives typically requires pre-calibration, like using checkerboards or MVS algorithms, which limits scalability and applicability in diverse real-world scenarios.
In this work, we present \textbf{\ours{}} (Reconstructing \textbf{A}rbitrary \textbf{H}umans from \textbf{A}rbitrary \textbf{P}erspectives), a feed-forward framework for reconstructing arbitrary humans from arbitrary camera perspectives without requiring camera calibration. Our core lies in the effective fusion of multi-view geometry to assist human association, reconstruction and localization.
Specifically, we use a Cross-View Identity Association module through learnable person queries and soft assignment, supervised by contrastive learning to resolve cross-view human identity association. A Human Head fuses cross-view features and scene context for SMPL prediction, guided by cross-view reprojection losses to enforce body pose consistency. Additionally, multi-view geometry eliminates the depth ambiguity inherent in monocular methods, providing more precise 3D human localization through multi-view triangulation.
Experiments on EgoHumans and EgoExo4D demonstrate that \ours{} achieves competitive performance on both world-space human reconstruction and camera pose estimation, while being 180$\times$ faster than optimization-based approaches.
\end{abstract}

\section{Introduction}
\label{sec:intro}

\begin{figure}[t]
    \centering
    \includegraphics[width=1.00\linewidth]{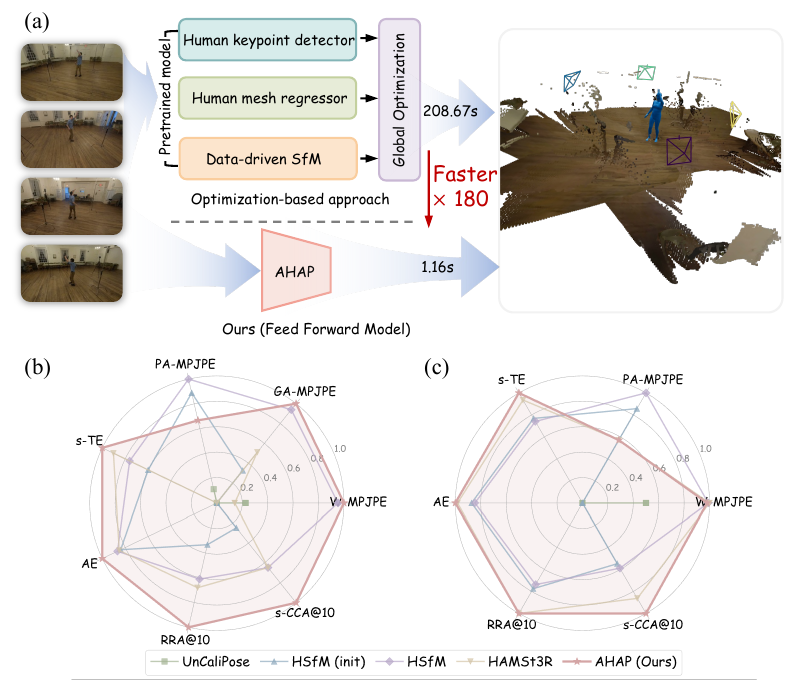}
    \caption{\textbf{(a)} \ours{} achieves 180$\times$ speedup over optimization-based HSfM~\cite{muller2025reconstructing} while maintaining competitive accuracy. \textbf{(b)} Results on EgoHumans. \textbf{(c)} Results on EgoExo4D.}
    \label{fig:abstract}
    \vspace{-15pt}
\end{figure}

Efficient and accurate reconstruction of both scene geometry~\cite{wang2024dust3r,wang2025vggt,lin2025depth}, and human bodies~\cite{ge20243d,wang2023zolly} in a shared world coordinate system is crucial for Embodied AI~\cite{wang2025sims,li2026w1,salzmann2023robots,fu2026egograsp,huang2025laytrol}, Virtual Reality, and Motion Analysis~\cite{dao2025beyond,chen2025language}. Multi-view observations~\cite{zhu2025viewmask} offer complementary visibility that alleviates occlusion and depth ambiguity while providing essential geometric constraints for aligning humans and scenes. Existing multi-view human-scene reconstruction methods~\cite{muller2025reconstructing,rojas2025hamst3r} typically rely on iterative test-time optimization to refine cross-view consistency among human meshes, cameras, and geometry. While accurate, these paradigms involve expensive optimization cycles and repeated reprojections , leading to high latency and limited scalability for real-time applications. This naturally raises the question:

\vspace{-3pt}
\begin{tcolorbox}[
    colback=gray!10,
    colframe=gray!10,
    boxrule=0pt,
    arc=4pt,
    left=6pt,
    right=6pt,
    top=4pt,
    bottom=4pt,
    boxsep=0pt
]
\textit{Can consistent 3D reconstruction of arbitrary humans and scenes be achieved in a purely feed-forward manner from arbitrary perspectives?}
\end{tcolorbox}
\vspace{-3pt}

Achieving this in uncalibrated, multi-person scenarios is particularly challenging. The same individual can exhibit drastic appearance changes and severe occlusions across viewpoints, while the co-occurrence of multiple people necessitates reliable cross-view identity association for correct fusion. Simultaneously, predictions from all viewpoints must be registered into a unified world coordinate system, requiring robust cross-view geometric reasoning and joint camera recovery without pre-calibrated extrinsics. Beyond the computational burden of optimization-based methods, current feed-forward alternatives fall short: single-view methods process viewpoints independently, failing to leverage cross-view complementarity~\cite{baradel2024multi,wang2025prompthmr}, while multi-view feed-forward methods either require pre-calibrated cameras or resort to post-hoc optimization for SMPL fitting, precluding end-to-end geometric supervision during training~\cite{muller2025reconstructing,rojas2025hamst3r}. Furthermore, many approaches remain restricted to single-person settings~\cite{li2024human}. Consequently, the joint recovery of multiple humans, scene geometry, and camera poses via a purely feed-forward pass from uncalibrated arbitrary-view inputs remains an open problem (\cref{tab:comparison}).

\begin{table}[t]
    \centering
    \caption{\textbf{Comparison of method capabilities.} \ours{} is the first feed-forward method to jointly reconstruct scene geometry and cameras from uncalibrated arbitrary-view images.}
    \vspace{-5pt}
    \label{tab:comparison}
    \setlength{\tabcolsep}{6pt}
    \renewcommand{\arraystretch}{1.00}
    \resizebox{\columnwidth}{!}{
        \begin{tabular}{l|cccc}
            \toprule
            \textbf{Method} & \textbf{Feed-fwd} & \textbf{Multi-Person} & \textbf{Camera} & \textbf{Scene} \\
            \midrule
            Multi-HMR~\cite{baradel2024multi} & \cmark & \cmark & \xmark & \xmark  \\
            SLAHMR~\cite{ye2023decoupling}  & \xmark & \cmark & \cmark & \xmark  \\
            UnCaliPose~\cite{xu2022multi}  & \xmark & \cmark & \cmark & \xmark  \\
            HSfM~\cite{muller2025reconstructing} & \xmark & \cmark & \cmark & \cmark \\
            HAMSt3R~\cite{rojas2025hamst3r} & \xmark & \cmark & \cmark & \cmark \\
            \midrule
            \rowcolor{best} \textbf{\ours{} (Ours)}  & \cmark & \cmark & \cmark & \cmark \\
            \bottomrule
        \end{tabular}
    }
    \vspace{-20pt}
\end{table}


To address these limitations, we propose \textbf{\ours{}} (Reconstructing \textbf{A}rbitrary \textbf{H}umans from \textbf{A}rbitrary \textbf{P}erspectives), a unified feed-forward framework for multi-person human-scene reconstruction from uncalibrated, arbitrary-view inputs. Our core insight lies in the effective fusion of multi-view geometry to jointly solve human association, reconstruction, and localization in a single forward pass. 
Specifically, we introduce a Cross-View Identity Association module that ensures robust identity consistency across disparate viewpoints via learnable person queries and a soft assignment mechanism. This learnable association is supervised by contrastive learning and leverages discriminative semantic features to resolve the challenges of appearance changes and severe occlusions, which effectively eliminates the necessity of traditional 2D tracking methods like~\cite{ye2023slahmr, rajasegaran2022tracking}. Guided by these associations, a specialized Human Head decodes aggregated multi-view features alongside dense scene context to directly regress SMPL parameters~\cite{loper2023smpl}. This joint reasoning significantly improves pose accuracy by leveraging complementary visibility to alleviate self-occlusions and depth ambiguities inherent in monocular methods~\cite{baradel2024multi, khirodkar2023ego}. The reconstruction is further regularized by a cross-view reprojection loss to enforce body pose consistency across all viewpoints. Finally, to overcome the imprecise 3D localization of single-view estimation, \ours{} leverages the precise 2D localization from our human branch and the 3D geometric reasoning of the scene branch to apply multi-view triangulation at inference. 

Experiments on EgoHumans~\cite{khirodkar2023ego} and EgoExo4D~\cite{grauman2024ego} demonstrate competitive performance on both human reconstruction and camera pose estimation, while achieving 180$\times$ speedup over optimization-based methods (\cref{fig:abstract}).
 Our main contributions are:
\begin{itemize}
    \item We present \ours{}, the first \textbf{feed-forward} framework for the joint reconstruction of arbitrary humans, scene geometry, and camera poses from uncalibrated arbitrary perspectives.
    \item We propose a cross-view \textbf{identity association} module that uses learnable queries and soft assignment to establish correspondences across viewpoints. This approach is supervised through contrastive learning to achieve discriminative matching, eliminating the reliance on visual tracking.
    \item We design a Human Head that jointly decodes cross-view aggregated features with scene context to predict SMPL parameters, supervised by cross-view reprojection losses for \textbf{body pose consistency}.
    \item We employ scale alignment and multi-view triangulation at inference to \textbf{localize human positions} and improve global human-scene alignment in the world coordinate system.    
\end{itemize}

\begin{figure*}[!t]
 \centering
\includegraphics[width=1.00\linewidth]{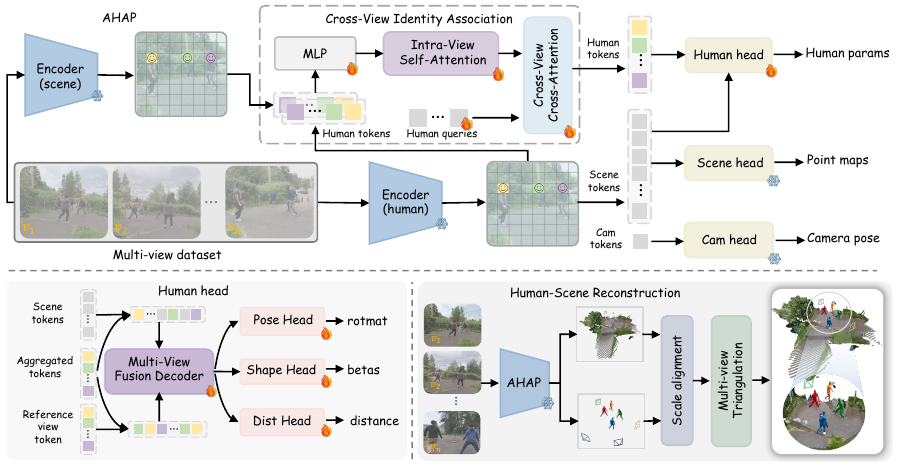}
\caption{
\textbf{Overall pipeline of \ours{}.}
Given multi-view images, the scene encoder~\cite{lin2025depth} estimates scene geometry and camera poses, while the human encoder~\cite{baradel2024multi} extracts human-centric features.
Our cross-view identity association module matches the same person across views via learnable queries. The human head fuses scene tokens, aggregated tokens, and reference view tokens through a multi-view fusion decoder to predict SMPL parameters. Finally, we align humans and scene point clouds via scale alignment and multi-view triangulation for precise human localization.
}
\label{fig:pipeline}
\vspace{-10pt}
\end{figure*}

\section{Related Works}
\label{sec:related}

\textbf{Multi-View Geometry.}
Classical Multi-View Geometry methods \cite{agarwal2011building,frahm2010building,liu2024robust}, such as COLMAP \cite{schonberger2016structure}, reconstruct 3D scenes via multi-stage pipelines with feature matching, triangulation, and bundle adjustment. While accurate, their reliance on handcrafted components and iterative optimization limits robustness and scalability. Recent differentiable SfM~\cite{wang2024vggsfm, lee2025dense} enables joint learning of geometry and camera parameters. In parallel, point-map-based approaches directly regress globally aligned 3D structures from sparse views, bypassing explicit optimization \cite{wang2024dust3r,leroy2024grounding}. VGGT \cite{wang2025vggt} unifies camera poses, depth, point maps, and tracks within a feed-forward transformer, while Depth Anything v3 (DA3)~\cite{lin2025depth} predicts spatially consistent 3D geometry and camera poses from arbitrary visual inputs via a unified depth-ray representation. Our method extends this paradigm by incorporating human reconstruction in a feed-forward manner, and further improves scene geometry prediction through human-scene joint reasoning.

\textbf{Feed-Forward Human Mesh Reconstruction.}
Feed-Forward Human mesh reconstruction aims to recover parametric body models, such as SMPL \cite{loper2023smpl} and SMPL-X \cite{pavlakos2019expressive}. Early methods focus on single-person monocular reconstruction by regressing body parameters from cropped images, with later approaches improving robustness through stronger backbones and large-scale supervision~\cite{wang2023zolly}. One-stage frameworks such as Multi-HMR \cite{baradel2024multi} extend to multi-person monocular reconstruction, while PromptHMR \cite{wang2025prompthmr} enables flexible control via spatial and semantic prompts on full images. Beyond monocular input, multi-view methods such as U-HMR~\cite{li2024human} exploit geometric constraints to resolve depth ambiguities, but remain limited to single-person scenarios. Multi-person multi-view reconstruction is still underexplored due to challenges in cross-view correspondence and inter-person occlusion, which motivates our feed-forward approach for arbitrary-human arbitrary-view reconstruction.

\textbf{Human-Scene Reconstruction.}
Joint reconstruction of humans and scenes has attracted increasing attention. HSfM~\cite{muller2025reconstructing} integrates human body constraints into a traditional Structure from Motion (SfM) pipeline to achieve metric-scale recovery, yet its reliance on iterative optimization limits scalability. HAMSt3R~\cite{rojas2025hamst3r} extends pointmap-based representation with human-specific heads for instance and DensePose prediction, though its post-hoc SMPL fitting remains a bottleneck. Human3R~\cite{chen2025human3r} achieves unified reconstruction from monocular video, it is constrained by single-view inputs. Earlier uncalibrated multi-view methods like UnCaliPose~\cite{xu2022multi} focus on pose estimation via cross-view matching but lack dense scene context. Our work operates in multi-view settings, enabling a unified feed-forward reconstruction of arbitrary humans and scenes from arbitrary perspectives.

\section{Method}
\label{sec:method}
Given $S$ multi-view  images ${I}_{i=1}^{S}$ of the same scene with human presence, our goal is to jointly reconstruct 3D human meshes and scene geometry with consistent identities across viewpoints in a feed-forward manner.
Unlike existing paradigms that either process views independently~\cite{baradel2024multi,wang2025prompthmr} or rely on expensive iterative optimization~\cite{xu2022multi,yin2025easyret3d,muller2025reconstructing}, \ours{} achieves this in a single forward pass.

\subsection{Overview}
\label{sec:overview}

As illustrated in \cref{fig:pipeline}, achieving arbitrary-view human-scene reconstruction requires addressing three core challenges: (1) establishing consistent human identity correspondences across uncalibrated views, (2) effectively aggregating multi-view features for accurate human mesh reconstruction, and (3) ensuring geometric consistency between reconstructed humans and the scene. To this end, \cref{sec:association} introduces a cross-attention-based association module that learns soft assignment across views. \cref{sec:smpl_prediction} presents the Human Head that fuses aggregated multi-view features with scene context for SMPL prediction. \cref{sec:supervision} details our multi-task supervision scheme enforcing cross-view geometric consistency. Finally, \cref{sec:triangulation} describes metric measurement and metric measurement and inference-time triangulation for global human-scene alignment.

\subsection{Visual Feature Encoding}
\label{sec:feature_extraction}

\begin{figure}[t]
    \centering
    \includegraphics[width=1.00\linewidth]{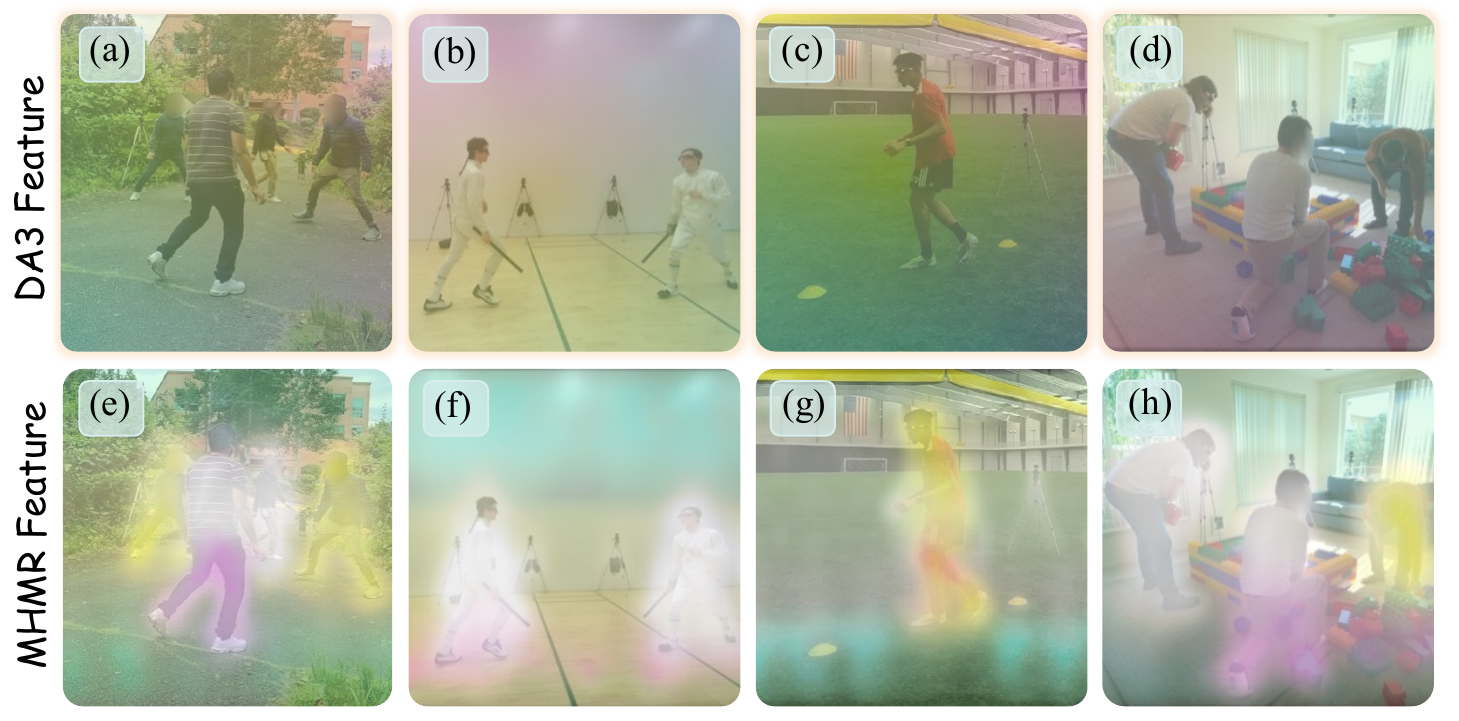}
    \caption{\textbf{PCA visualization of feature distributions.} (a-d) Scene encoder~\cite{lin2025depth} features; (e-h) Human encoder~\cite{baradel2024multi} features. Human encoder features show stronger semantic clustering aligned with ground truth, while scene encoder provides complementary geometric information.}
    \label{fig:pca}
    \vspace{-10pt}    
\end{figure}

To capture complementary visual information, we leverage two pre-trained backbones.
The scene branch utilizes DA3~\cite{lin2025depth} to extract dense geometric features $\bm{F}^{\text{scene}}$, while the human branch employs a Multi-HMR~\cite{baradel2024multi} backbone to extract human-centric semantic features $\bm{F}^{\text{human}}$ alongside a detection head for localization.
As shown in \cref{fig:pca}, human features exhibit strong semantic clustering, whereas scene encoder provides critical geometric context. At each detected human location $(h,w)$ in the human feature map, we map to the corresponding scene location $(h',w')$ via resolution alignment, then fuse tokens through a two-layer MLP
\begin{equation}
\mathbf{z} = \mathcal{M}_{\text{fuse}} \left( \big[ \bm{F}_{h,w}^{\text{human}} \,;\, \bm{F}_{h',w'}^{\text{scene}} \big] \right) \in \mathbb{R}^{D},
\end{equation}
where $D=1024$. These fused tokens integrate discriminative appearance cues with scene-aware geometric information, serving as the input for our cross-view identity association module.

\subsection{Cross-View Identity Association}
\label{sec:association}

The core challenge in multi-view human reconstruction is establishing human identity correspondences across views without explicit 3D supervision. We propose a soft-assignment based association module that learns to match detections belonging to the same person.

Before cross-view association, we first enhance each detection representation through intra-view self-attention.
For detections $\{\bm{z}_1, \ldots, \bm{z}_{N_v}\}$ in view $v$, multi-head self-attention is applied to enable awareness of other people in the same view
\begin{equation}
\tilde{\bm{z}}_j = \mathrm{LN} \left( \bm{z}_j + \mathrm{SA} \big( \bm{z}_j, \{ \bm{z}_k \}_{k=1}^{N_v} \big) \right),
\end{equation}

To establish cross-view correspondences, learnable person queries $\bm{Q} \in \mathbb{R}^{P \times D}$ are introduced, where $P=36$ is the maximum number of persons and $D = 1024$. Given all enhanced detection tokens $\{\tilde{\bm{z}}_n\}_{n=1}^{N}$ from all views, we compute soft assignment through multi-head cross-attention
\begin{equation}
\bm{A} = \frac{1}{H}\sum_{h=1}^{H}\text{softmax}\left(\frac{\bm{Q}\bm{W}_h^Q (\bm{K}\bm{W}_h^K)^\top}{\sqrt{D/H}}\right) \in \bm{R}^{P \times N},
\label{eq:soft_assignment}
\end{equation}
where $\bm{K} = [\tilde{\bm{z}}_1, \ldots, \tilde{\bm{z}}_N]^\top$ are the detection tokens serving as keys and values, and $H=8$ is the number of attention heads.
Each row $\bm{A}_{p,:}$ represents the probability distribution of person query $p$ attending to all detections, effectively learning which detections belong to the same individual.
The final assignment is $\text{person}(n) = \arg\max_{p} \bm{A}_{p,n}$.

For each person $p$, we aggregate features from all associated detections via weighted sum
\begin{equation}
\vspace{-5pt}
\bm{f}_p^{\text{agg}} = \text{LN}\left(\bm{A}_{p,:} \cdot \bm{K}\right) \in \mathbb{R}^{D}.
\end{equation}
This weighted sum naturally handles occlusions and varying numbers of viewpoints per person by prioritizing high-confidence features.

\subsection{Arbitrary View Human Head}
\label{sec:smpl_prediction}

To achieve cross-view consistent SMPL recovery, the Human Head incorporates aggregated multi-view information into a transformer-based decoder. While prediction is performed within a reference view, typically the one with the highest detection density, the process is guided by the global representation $\bm{f}_p^{\text{agg}}$.

We construct context features by combining human features $\bm{F}^{\text{human}} \in \mathbb{R}^{H \times W \times 1024}$, projected DA3 features $\text{Proj}(\bm{F}^{\text{scene}}) \in \mathbb{R}^{H \times W \times 1024}$, and Fourier camera ray embeddings $\bm{E}^{\text{cam}} \in \bm{R}^{H \times W \times 99}$~\cite{mildenhall2021nerf}.
The aggregated multi-view feature $\bm{f}_p^{\text{agg}}$ is then projected and added to the context
\begin{equation}
\tilde{\bm{C}} = \bm{C} + \sigma(\alpha) \cdot \text{Broadcast}\left(\text{Proj}_{\text{kv}}(\bm{f}_p^{\text{agg}})\right),
\end{equation}
where $\sigma(\alpha)$ is a learnable scale initialized near zero for training stability.
Additionally, the aggregated feature is projected as an auxiliary query token $\bm{q}_{\text{mv}} = \text{Proj}_{\text{q}}(\bm{f}_p^{\text{agg}})$.

The context feature at the detection location is augmented with mean SMPL parameters to form the primary query $\bm{q}_{\text{det}} = [\bm{z}; \bar{\boldsymbol{\theta}}; \bar{\boldsymbol{\beta}}; \bar{\bm{c}}]$.
Both queries are processed together through cross-attention~\cite{jaegle2021perceiver,dosovitskiy2020image} with the multi-view enhanced context:
\begin{equation}
\bm{q}_{\text{final}} = \text{CA}\left([\bm{q}_{\text{det}}; \bm{q}_{\text{mv}}], \tilde{\bm{C}}, \tilde{\bm{C}}\right).
\end{equation}
The final prediction is
\begin{align}
\boldsymbol{\theta} &= \bar{\boldsymbol{\theta}} + \mathcal{M}_{\text{pose}}(\bm{q}_{\text{final}}), \\
\boldsymbol{\beta} &= \bar{\boldsymbol{\beta}} + \mathcal{M}_{\text{shape}}(\bm{q}_{\text{final}}), \\
\bm{d} &= \mathcal{M}_{\text{cam}}(\bm{q}_{\text{final}}),
\end{align}
where $\boldsymbol{\theta}$ denotes pose parameters, $\boldsymbol{\beta}$ denotes shape parameters, and $\bm{d}$ denotes distance~\cite{kanazawa2018end}. Unlike post-hoc fitting methods that rely on iterative refinement, this feed-forward head directly leverages multi-view complementarity to resolve depth and pose ambiguities.

\begin{figure}[t]
    \centering
    \includegraphics[width=1.00\linewidth]{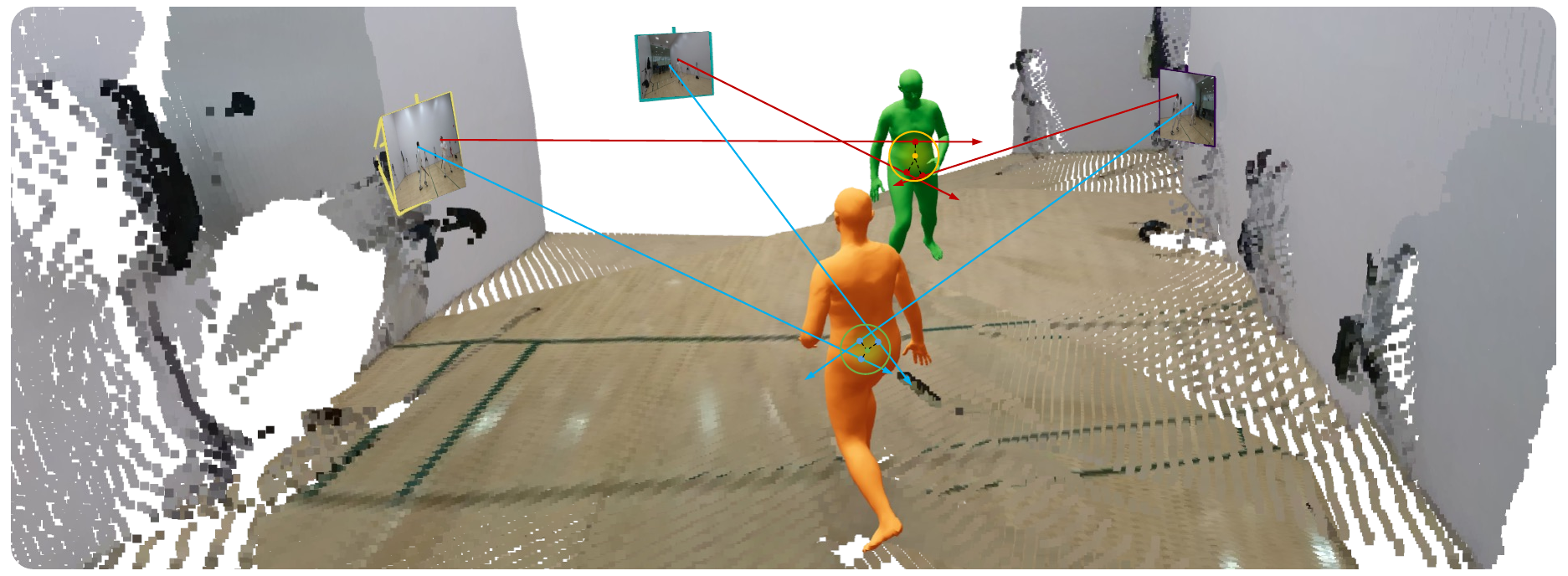}
    \caption{\textbf{Multi-view triangulation for human position refinement.} For persons visible in multiple views, we refine their 3D positions using DLT triangulation based on 2D pelvis observations and estimated camera poses, improving human-scene alignment.}
    \label{fig:triangulation}
    \vspace{-10pt}    
\end{figure}

\subsection{Training Details}
\label{sec:supervision}

To supervise the association module and enforce geometric consistency, three complementary loss functions are designed.

\noindent\textbf{Soft Assignment Loss.}
We directly supervise the attention weights in \cref{eq:soft_assignment} using ground-truth correspondences.
Given the target assignment matrix $\bm{A}^*$, predicted identities are aligned with ground-truth labels via Hungarian matching, followed by the computation of a cross-entropy loss:
\begin{equation}
\mathcal{L}_{\text{assign}} = -\frac{1}{P^*} \sum_{p=1}^{P^*} \sum_{n=1}^{N} \tilde{\bm{A}}^*_{p,n} \log \bm{A}_{\pi(p),n},
\end{equation}
where $\pi$ is the optimal matching and $\tilde{\bm{A}}^*$ is row-normalized.

\noindent\textbf{Contrastive Loss.}
To learn discriminative features for association, we apply InfoNCE~\cite{oord2018representation} where detections of the same person across views form positive pairs
\begin{equation}
\mathcal{L}_{\text{contra}} = -\frac{1}{|\mathcal{P}|} \sum_{(i,j) \in \mathcal{P}} \log \frac{\exp(\bm{h}_i \cdot \bm{h}_j^+ / \tau)}{\sum_{k} \exp(\bm{h}_i \cdot \bm{h}_k / \tau)},
\end{equation}
where $\tau = 0.07$ is the temperature and $\bm{h} = \text{Proj}(\tilde{\bm{z}})$ are L2-normalized projections of the enhanced human tokens.

\noindent\textbf{Cross-View Reprojection Loss.}
To enforce geometric consistency, we transform predictions from the reference view to other views using camera extrinsics ($\bm{J}_{\text{world}} = \bm{C2W}^{\text{ref}} \cdot \bm{J}^{\text{ref}}_{\text{cam}}$, then $\bm{J}^{v}_{\text{cam}} = \bm{W2C}^{v} \cdot \bm{J}_{\text{world}}$) and compare with GT:
\begin{equation}
\mathcal{L}_{\text{reproj}} = \frac{1}{|\mathcal{V}|} \sum_{v \in \mathcal{V}} \left( \|\bm{J}^{v}_{\text{rel}} - \hat{\bm{J}}^{v}_{\text{rel}}\|_1 + \lambda \|\Pi(\bm{J}^{v}) - \hat{\bm{j}}^{v}_{2d}\|_1 \right),
\end{equation}
where $\bm{J}_{\text{rel}}$ denotes root-relative 3D joints and $\Pi(\cdot)$ is perspective projection.

The complete training objective is
\begin{equation}
\mathcal{L} = \mathcal{L}_{\text{SMPL}} + \lambda_1 \mathcal{L}_{\text{assign}} + \lambda_2 \mathcal{L}_{\text{contra}} + \lambda_3 \mathcal{L}_{\text{reproj}},
\end{equation}
where $\mathcal{L}_{\text{SMPL}}$ is the standard SMPL supervision loss (details in supplementary material).

\noindent\textbf{Two-Stage Training Strategy.}
We adopt a two-stage curriculum: first traininging on BEDLAM~\cite{black2023bedlam} temporal sequences. Consecutive frames serve as pseudo multi-view inputs with minimal appearance variation, allowing the association module to learn robust identity matching in a simplified setting. Then fine-tuning on EgoHumans~\cite{khirodkar2023ego} multi-view data. This stage enables the model to generalize to real cross-view scenarios characterized by severe occlusions and drastic appearance changes.

\begin{table*}[ht]
  \caption{\textbf{Human-centric evaluation metrics} on EgoHumans and EgoExo4D. \ours{} achieves competitive W-MPJPE (0.88m) and GA-MPJPE (0.18m) on EgoHumans, outperforming optimization-based HSfM by 15\%. While HSfM attains lower PA-MPJPE through iterative refinement, our feed-forward approach remains competitive (0.08m vs. 0.05m) without test-time optimization. \highlightnote}
  \vspace{-0.2cm}
  \centering
  \setlength{\tabcolsep}{8pt}
  \renewcommand{\arraystretch}{1.20}
  \resizebox{0.80\textwidth}{!}{%
    \begin{tabular}{l|l| ccc|cc}
      \toprule
      \multirow{2.5}{*}{\textbf{Type}} & \multirow{2.5}{*}{\textbf{Method}} & \multicolumn{3}{c|}{\textbf{EgoHumans}} & \multicolumn{2}{c}{\textbf{EgoExo4D}} \\
      \cmidrule(lr){3-5} \cmidrule(lr){6-7}
      & & W-MPJPE$\downarrow$ & GA-MPJPE$\downarrow$ & PA-MPJPE$\downarrow$ & W-MPJPE$\downarrow$ & PA-MPJPE$\downarrow$ \\
      \midrule
      \textit{Single-view} & Multi-HMR~\cite{baradel2024multi} & 7.66 & 0.99 & 0.12 & 2.88 & \cellcolor{second}\underline{0.07} \\
      \midrule
      \multirow{4}{*}{\textit{Optimization-based}} & UnCaliPose~\cite{xu2022multi} & 3.51 & 0.67 & 0.13 & 2.90 & 0.13 \\
      & HSfM (init)~\cite{muller2025reconstructing} & 4.28 & 0.51 & \cellcolor{second}\underline{0.06} & 5.29 & \cellcolor{second}\underline{0.07} \\
      & HSfM~\cite{muller2025reconstructing} & \cellcolor{second}\underline{1.04} & \cellcolor{second}\underline{0.21} & \cellcolor{best}\textbf{0.05} & \cellcolor{second}\underline{0.56} & \cellcolor{best}\textbf{0.06} \\
      & HAMSt3R~\cite{rojas2025hamst3r} & 3.80 & 0.42 & 0.14 & \cellcolor{best}\textbf{0.51} & 0.09 \\
      \midrule
      \textit{\textbf{Feed-forward}} & \textbf{\ours{} (Ours)} & \cellcolor{best}\textbf{0.88} & \cellcolor{best}\textbf{0.18} & 0.08 & 0.60 & 0.09 \\
      \bottomrule
    \end{tabular}%
  }
  \label{tab:human_metrics}
\end{table*}

\begin{table*}[ht]
    \centering
    \caption{\textbf{Camera pose evaluation} on EgoHumans and EgoExo4D. Feed-forward methods achieve superior rotation accuracy: \ours{} obtains AE of 1.47° on EgoHumans (vs. HSfM's 9.35°) with perfect RRA@10 of 1.00. Compared to DA3, \ours{} improves rotation accuracy by 35\% (AE: 1.47° vs. 2.26°) and scale-aligned translation by 41\% (s-TE: 0.16m vs. 0.27m), demonstrating that joint human-scene learning provides mutually beneficial geometric priors. \highlightnote}
    \vspace{-0.2cm}
    \setlength{\tabcolsep}{6pt}
    \renewcommand{\arraystretch}{1.00}
    \resizebox{\textwidth}{!}{%
        \begin{tabular}{c|l| cccccc|cccccc}
        \toprule
        \multirow{2.5}{*}{\textbf{Type}} & \multirow{2.5}{*}{\textbf{Method}}
            & \multicolumn{6}{c|}{\textbf{EgoHumans}}
            & \multicolumn{6}{c}{\textbf{EgoExo4D}} \\
        \cmidrule(lr){3-8} \cmidrule(lr){9-14}
        & & TE$\downarrow$ & s-TE$\downarrow$ & AE$\downarrow$ & RRA@10$\uparrow$ & CCA@10$\uparrow$ & s-CCA@10$\uparrow$
        & TE$\downarrow$ & s-TE$\downarrow$ & AE$\downarrow$ & RRA@10$\uparrow$ & CCA@10$\uparrow$ & s-CCA@10$\uparrow$ \\
        \midrule
        \multirow{4}{*}{\rotatebox{90}{\textit{Opt.-based}}} & UnCaliPose~\cite{xu2022multi} & 2.63 & 2.63 & 60.90 & 0.28 & - & 0.33 & 2.43 & 1.16 & 65.61 & 0.19 & - & 0.24 \\
        & HSfM (init)~\cite{muller2025reconstructing} & 2.37 & 1.15 & 11.00 & 0.52 & \cellcolor{second}\underline{0.26} & 0.49 & 1.27 & 0.33 & 9.92 & 0.81 & 0.05 & 0.64 \\
        & HSfM~\cite{muller2025reconstructing} & \cellcolor{best}\textbf{2.09} & 0.75 & 9.35 & 0.72 & \cellcolor{best}\textbf{0.32} & 0.75 & \cellcolor{second}\underline{0.95} & 0.36 & 11.57 & 0.78 & 0.07 & 0.67 \\
        & HAMSt3R~\cite{rojas2025hamst3r} & \cellcolor{second}\underline{2.33} & 0.40 & 10.24 & 0.77 & 0.06 & 0.75 & \cellcolor{best}\textbf{0.60} & 0.15 & 2.85 & \cellcolor{best}\textbf{0.99} & \cellcolor{best}\textbf{0.42} & 0.87 \\
        \midrule
        \multirow{2}{*}{\rotatebox{90}{\textit{\textbf{FF}}}} & DA3~\cite{lin2025depth} & 3.89 & \cellcolor{second}\underline{0.27} & \cellcolor{second}\underline{2.26} & \cellcolor{second}\underline{0.99} & - & \cellcolor{second}\underline{0.86} & 1.62 &\cellcolor{best} \textbf{0.07} & \cellcolor{second}\underline{1.80} & \cellcolor{second}\underline{0.98} & 0.02 & \cellcolor{second}\underline{0.96} \\
        & \textbf{\ours{} (Ours)} & 4.54 & \cellcolor{best}\textbf{0.16} & \cellcolor{best}\textbf{1.47} & \cellcolor{best}\textbf{1.00} & - & \cellcolor{best}\textbf{0.98} & 1.60 &\cellcolor{second} \underline{0.08} & \cellcolor{best}\textbf{1.74} & \cellcolor{best}\textbf{0.99} & \cellcolor{second}\underline{0.30} & \cellcolor{best}\textbf{0.97} \\
        \bottomrule
        \end{tabular}%
    }%
    \label{tab:camera_metrics}
    \vspace{-10pt}
\end{table*}

\begin{table}[t]
    \centering
    \caption{\textbf{Efficiency comparison.} Our method achieves 180--250$\times$ speedup over HSfM across different numbers of views.}
    \vspace{-0.2cm}
    \setlength{\tabcolsep}{8pt}
    \renewcommand{\arraystretch}{1.20}
    \resizebox{\columnwidth}{!}{%
    \begin{tabular}{l| cccc}
        \toprule
        \multirow{2}{*}{\textbf{Method}} & \multicolumn{4}{c}{\textbf{Num. Views}} \\
        \cmidrule(lr){2-5}
        & 1 & 2 & 3 & 4 \\
        \midrule
        HSfM~\cite{muller2025reconstructing} & - & 180.41s & 201.85s & 208.67s \\
        \textbf{\ours{} (Ours)} & \cellcolor{best}\textbf{0.34s} & \cellcolor{best}\textbf{0.72s} {\scriptsize(250$\times$)} & \cellcolor{best}\textbf{0.94s} {\scriptsize(215$\times$)} & \cellcolor{best}\textbf{1.16s} {\scriptsize(180$\times$)} \\
        \bottomrule
    \end{tabular}%
    }
    \label{tab:efficiency}
    \vspace{-10pt}      
\end{table}

\subsection{Human-Scene Reconstruction}
\label{sec:triangulation}

During inference, we utilize multi-view geometry to refine human positions and improve the overall human-scene alignment.
The camera poses are estimated by the scene branch, which predicts camera-to-world transforms $\{\bm{T}^{c2w}_i\}$ relative to a reference view.

Since SMPL predictions and scene depth may have inconsistent scales, we first compute a global scale factor.
For each detected human, we sample the scene depth at the pelvis 2D location using a local window and take the median, then compare with the predicted pelvis depth
\begin{equation}
s = \text{median}\left(\frac{z^{\text{scene}}_i}{z^{\text{human}}_i}\right)_{i=1}^{N},
\end{equation}
where $N$ is the number of humans with valid depth pairs.
This global scale is applied to all human meshes to ensure consistent sizing across the scene.

For individuals visible in $\geq 2$ views, we further refine their 3D positions through Direct Linear Transform (DLT) triangulation.
Given 2D pelvis observations $\{(u_i, v_i)\}$ and projection matrices $\bm{P}_i = \bm{K}_i [\bm{R}_i | \bm{t}_i]$ derived from the estimated camera poses, we construct the linear system:
\begin{align}
\bm{A}\bm{X} &= \bm{0}, \quad \text{where} \\
\bm{A}_{2i} &= u_i \bm{P}_i^{(3)} - \bm{P}_i^{(1)}, \quad \bm{A}_{2i+1} = v_i \bm{P}_i^{(3)} - \bm{P}_i^{(2)}. \nonumber
\end{align}
The solution is obtained via SVD as $\bm{X}^* = \arg\min_{\|\bm{X}\|=1} \|\bm{A}\bm{X}\|$, where $\bm{A}$ is constructed from the 2D pelvis projections across views.
We further transform the triangulated translation to camera space for 2D overlay.
This refinement step is a closed-form geometric solution that maintains high computational efficiency.

\begin{figure*}[!t]
    \centering
    \includegraphics[width=1.00\textwidth]{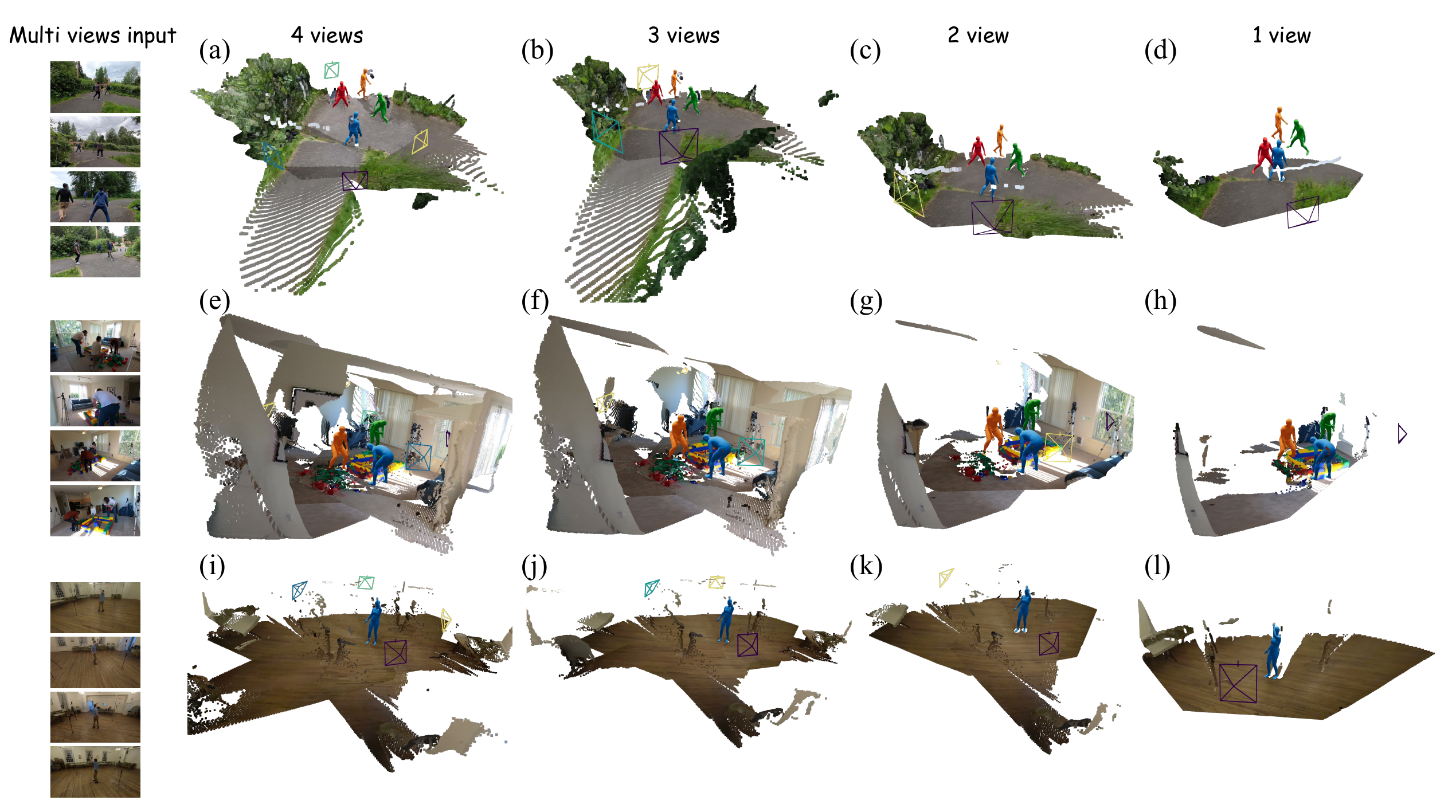}
    \vspace{-5pt}
    \caption{\textbf{Qualitative results.} Visualization of human-scene reconstruction on EgoHumans and EgoExo4D. AHAP produces accurate human meshes within reconstructed scenes, maintaining consistent identity association across views.}
    \label{fig:case}
    \includegraphics[width=0.93\textwidth]{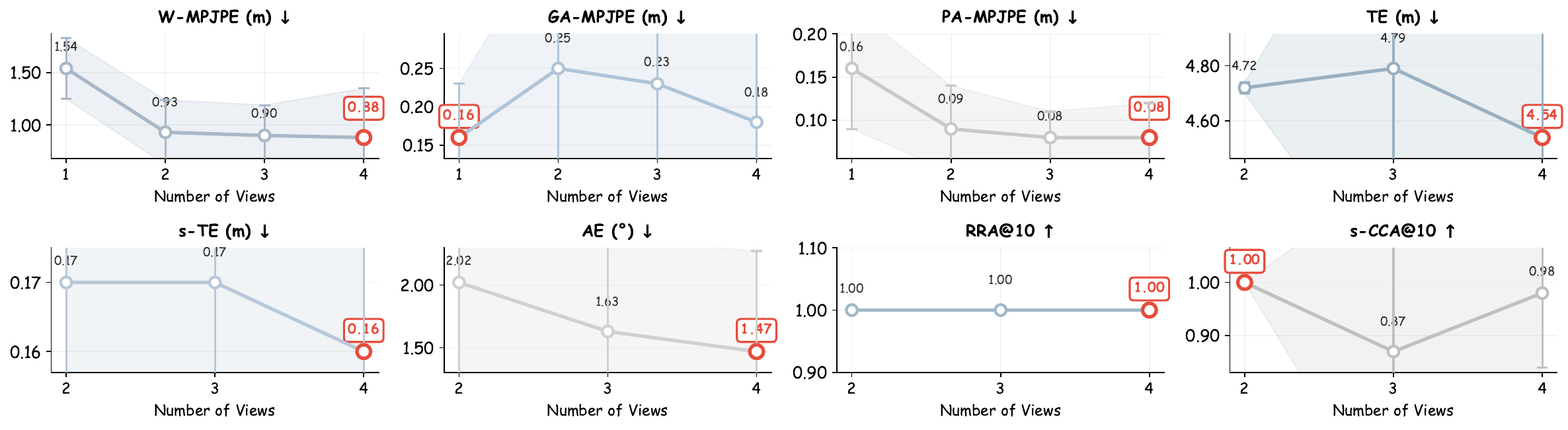}
    \vspace{-5pt}
    \caption{\textbf{Ablation study on the number of input views} on EgoHumans. \ours{} consistently improves as more views become available. The first additional view provides the largest gain (W-MPJPE: 1.54m $\rightarrow$ 0.93m), with continued improvements at 3--4 views. Camera pose accuracy also benefits, with angular error reducing from 2.02° to 1.47°.}
    \vspace{-15pt}
    \label{fig:numviews_ablation}
\end{figure*}

\section{Experiments}

\subsection{Datasets and Metrics}

AHAP is trained on the training sets of BEDLAM~\cite{black2023bedlam} and EgoHumans~\cite{khirodkar2023ego}. Generalization is evaluated on the test splits of EgoHumans and EgoExo4D~\cite{grauman2024ego}. For human mesh recovery, we report W-MPJPE, GA-MPJPE, and PA-MPJPE to measure 3D joint errors across different alignment levels. Camera pose accuracy is evaluated using TE, s-TE, AE, RRA@10~\cite{wang2023posediffusion}, CCA@10~\cite{lin2024relpose++}, and s-CCA@10~\cite{muller2025reconstructing}. These metrics assess both metric-scale and scale-invariant translation and rotation performance. Detailed definitions are provided in the supplementary material.

\subsection{Results}

\ours{} is evaluated against several baseline methods: the monocular Multi-HMR~\cite{baradel2024multi}, the uncalibrated pose estimator UnCaliPose~\cite{xu2022multi}, the optimization-based HSfM~\cite{muller2025reconstructing}, and the hybrid framework HAMSt3R~\cite{rojas2025hamst3r}. Quantitative comparisons on the EgoHumans and EgoExo4D datasets are summarized in \cref{tab:human_metrics,tab:camera_metrics}. Across both benchmarks, \ours{} achieves competitive performance in world-space human reconstruction and camera pose estimation. Unlike optimization-based HSfM and HAMSt3R, \ours{} recovers SMPL parameters and camera extrinsics via a strictly feed-forward inference. This architecture offers a superior balance between reconstruction fidelity and computational efficiency, yielding a 180$\times$ speedup over optimization-heavy baselines.
\begin{figure*}[h]
 \centering
\includegraphics[width=0.93\linewidth]{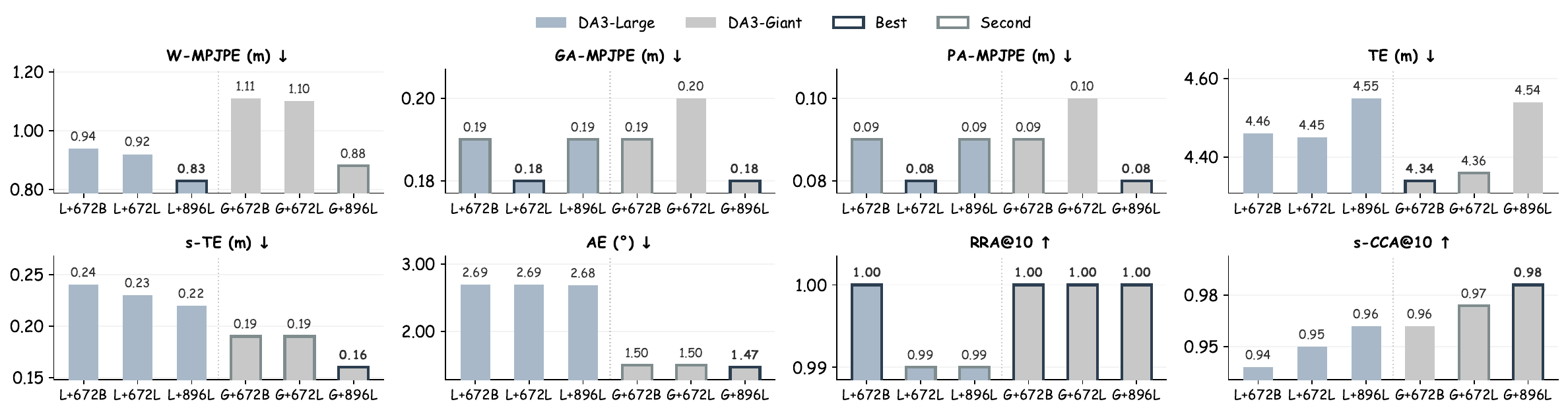}
\vspace{-5pt}
\caption{
\textbf{Ablation study on backbone variants.} We evaluate combinations of DA3 (Large/Giant) and Multi-HMR (672B/672L/896L) backbones. Giant variants excel in camera metrics while 896L improves human reconstruction.
}
\label{fig:backbone_ablation}
\vspace{-10pt}
\end{figure*}
\paragraph{Human Mesh Recovery.}
As shown in \cref{tab:human_metrics}, \ours{} achieves competitive performance on EgoHumans, particularly excelling in world-space metrics (W-MPJPE, GA-MPJPE). This advantage stems from our cross-view association module, which effectively resolves the inherent depth ambiguity in monocular estimation by aggregating complementary observations. The significant performance gap between multi-view methods and monocular Multi-HMR validates the necessity of multi-view reasoning. While optimization-based HSfM achieves lower PA-MPJPE through iterative pose refinement, our feed-forward approach demonstrates stronger global positioning capability without test-time optimization.
\vspace{-10pt}
\paragraph{Camera Pose Estimation.}
As shown in \cref{tab:camera_metrics}, feed-forward methods demonstrate clear advantages in rotation accuracy over optimization-based approaches, likely because learned priors are more robust to local minima than iterative optimization. Notably, \ours{} consistently outperforms the DA3 baseline across rotation and scale-aligned translation metrics, validating that human bodies serve as effective semantic landmarks providing complementary geometric constraints beyond scene features alone. While our absolute translation error is higher, the excellent scale-aligned performance indicates that the primary source of error is scale ambiguity rather than geometric misalignment---a common challenge in monocular-initialized systems that does not affect downstream human-scene consistency.
\vspace{-10pt}
\paragraph{Efficiency Analysis.}
Unlike optimization-based approaches that necessitate costly iterative refinement, \ours{} performs reconstruction in a single forward pass. Inference latency was benchmarked on an NVIDIA H200 GPU using 4-view inputs. While HSfM requires approximately 209 seconds for iterative optimization, AHAP completes the same task in only 1.16 seconds, representing a 180$\times$ speedup. Furthermore, our framework demonstrates excellent scalability, with execution time increasing only marginally from 0.89s to 1.16s when scaling from 2 to 4 views.
\vspace{-10pt}
\paragraph{Qualitative Results.}
Visualizations in \cref{fig:case} demonstrate that AHAP produces accurate human meshes with consistent identity association. While monocular estimations suffer from severe depth ambiguity and poor localization, our multi-view framework leverages cross-view triangulation and geometric supervision to correctly place humans in world coordinates. These mechanisms effectively enforce spatial consistency across disparate vantage points. A detailed quantitative analysis of how performance scales with the number of input views is provided in the ablation studies.
\vspace{-20pt}
\subsection{Ablation Studies}
\paragraph{Impact of Input Views.}
We evaluate the sensitivity of \ours{} to the number of input views during inference (\cref{fig:numviews_ablation}). Single-view reconstruction yields a W-MPJPE of 1.54m and a PA-MPJPE of 0.16m, reflecting inherent monocular depth ambiguities. The introduction of a second view significantly improves these metrics to 0.93m and 0.09m, respectively, before performance gains begin to plateau at 3--4 views. Regarding extrinsic calibration, translation error (TE) remains stable (4.5--4.8m), while the scale-corrected translation error (s-TE) is consistently low ($\leq$0.17m), confirming the accurate recovery of relative camera geometry. Furthermore, angular error (AE) improves steadily from 2.02° to 1.47° as additional views are incorporated. Robustness is further evidenced by registration metrics, with RRA@10 remaining at 1.00 and s-CCA@10 $\geq$0.87 across all multi-view configurations. 

\vspace{-10pt}
\paragraph{Backbone Variants.}
We evaluate the impact of different backbone scales by combining two DA3~\cite{lin2025depth} variants (Giant and Large) with three Multi-HMR~\cite{baradel2024multi} variants (896L, 672L, and 672B). As shown in \cref{fig:backbone_ablation}, the Giant+896L configuration achieves the best overall performance. For camera pose estimation, DA3 backbone scale is the dominant factor: Giant variants consistently outperform Large variants in rotation accuracy (AE: 1.47° vs. 2.68°) and translation precision (s-TE: 0.16m vs. 0.22m), indicating that stronger geometric priors from the scene branch are essential for accurate camera recovery. For human reconstruction, Multi-HMR resolution plays a more significant role: the 896L variant yields the best W-MPJPE across both DA3 scales (0.83m for Large, 0.88m for Giant), suggesting that higher-resolution human features improve localization accuracy. We adopt Giant+896L as our default configuration to achieve the best balance between human reconstruction and camera estimation.
\begin{table}[t]
    \centering
    \caption{\textbf{Ablation study on key components} on EgoHumans.}
    \vspace{-0.2cm}
    \renewcommand{\arraystretch}{1.00}
    \resizebox{0.50\textwidth}{!}{%
    \begin{tabular}{l|ccc}
        \toprule
        \textbf{Method} & W-MPJPE$\downarrow$ & GA-MPJPE$\downarrow$ & PA-MPJPE$\downarrow$ \\
        \midrule
        w/o $\mathcal{L}_{\text{reproj}}$ & 1.33 & 0.25 & 0.12 \\
        w/o $\mathcal{L}_{\text{assign}}$ & 1.33 & 0.24 & 0.11 \\
        w/o $\mathcal{L}_{\text{contra}}$ & 1.30 & 0.23 & 0.12 \\
        w/o $\mathcal{L}_{\text{assign + contra}}$ & 1.47 & 0.39 & 0.14 \\
        \midrule
        w/o triangulation & 0.91 & 0.19 & 0.08 \\
        \midrule
        \rowcolor{gray!15}
        \textbf{Full Model} & \textbf{0.88} & \textbf{0.18} & \textbf{0.08} \\
        \bottomrule
    \end{tabular}%
    }
    \label{tab:ablation_loss}
    \vspace{-15pt}    
\end{table}

\paragraph{Key Component Ablation.}
We investigate the contribution of each component in \cref{tab:ablation_loss}. For loss functions, removing any term leads to performance degradation, confirming their synergistic roles. The reprojection loss $\mathcal{L}_{\text{reproj}}$ is the most critical; its absence results in a 51\% increase in W-MPJPE (0.88m to 1.33m), as the model loses the geometric constraints necessary for cross-view consistency. The assignment loss $\mathcal{L}_{\text{assign}}$ and contrastive loss $\mathcal{L}_{\text{contra}}$ are equally vital for identity association; removing both simultaneously causes GA-MPJPE to more than double (0.18m to 0.39m). For the triangulation module, removing it leads to a modest but consistent degradation across all metrics (W-MPJPE: 0.88m to 0.91m), demonstrating that explicit 3D reasoning via triangulation provides complementary geometric cues beyond the learned features alone.

\vspace{-10pt}

\section{Conclusion}

We presented \ours{}, a feed-forward framework for multi-view multi-person human reconstruction from uncalibrated cameras. Our method addresses three key challenges simultaneously: cross-view identity association through learnable person queries with soft assignment, geometric consistency via association-guided SMPL decoding, and camera-free operation by jointly estimating scene geometry and camera poses. The multi-view supervision with contrastive learning and cross-view reprojection losses enables end-to-end training with strong geometric constraints. Experiments on EgoHumans and EgoExo4D demonstrate that \ours{} achieves competitive performance while being 180$\times$ faster than optimization-based methods, making it practical for real-world applications.

\section*{Acknowledgments}
This paper is partially supported by the National Key R\&D Program of China No.2022ZD0161000.

\section*{Impact Statements}
This paper presents work whose goal is to advance the field of machine learning. There are many potential societal consequences of our work, none of which we feel must be specifically highlighted here.


\bibliography{reference}
\bibliographystyle{icml2026}

\newpage
\appendix
\onecolumn
\section{Appendix.}

\section{Technical Details}
\label{sec:suppl_technical}

\subsection{SMPL Supervision Loss}
\label{sec:suppl_smpl_loss}

We directly supervise SMPL parameters in their respective representation spaces:

\begin{equation}
\mathcal{L}_{\text{SMPL}} = \lambda_{\text{det}} \mathcal{L}_{\text{det}} + \lambda_{\text{param}} \mathcal{L}_{\text{param}} + \lambda_{\text{3D}} \mathcal{L}_{\text{3D}} + \lambda_{\text{2D}} \mathcal{L}_{\text{2D}}.
\end{equation}

\noindent\textbf{Detection Loss.}
We employ a modified focal loss This loss addresses class imbalance by down-weighting easy samples:
\begin{equation}
\mathcal{L}_{\text{det}} = -\frac{1}{N_{\text{pos}}} \sum_{i} \begin{cases}
(1-\hat{p}_i)^2 \log(\hat{p}_i) & \text{if } p_i = 1 \\
(1-p_i)^4 \hat{p}_i^2 \log(1-\hat{p}_i) & \text{otherwise}
\end{cases},
\end{equation}
where $p_i$ and $\hat{p}_i$ denote the ground-truth and predicted scores at location $i$, respectively. Gaussian smoothing is applied to $p_i$ around human centers. To achieve sub-pixel localization accuracy, we further supervise the 2D offset $\Delta \mathbf{u}$ from patch centers:
\begin{equation}
\mathcal{L}_{\text{offset}} = \|\Delta \hat{\mathbf{u}} - \Delta \mathbf{u}\|_1.
\end{equation}

\noindent\textbf{SMPL Parameter Loss.}
We directly supervise SMPL parameters in their respective representation spacess~\cite{goel2023humans,kolotouros2019learning}:
\begin{align}
\mathcal{L}_{\text{param}} &= \lambda_R\|\hat{\mathbf{R}} - \mathbf{R}\|_1 + \lambda_\beta\|\hat{\boldsymbol{\beta}} - \boldsymbol{\beta}\|_1 + \lambda_d|\hat{d} - d| + \lambda_t\|\hat{\mathbf{t}} - \mathbf{t}\|_1,
\end{align}
where $\mathbf{R} \in \mathbb{R}^{24 \times 3 \times 3}$ represents joint $SO(3)$ to ensgures (including global orientation), supervised in $SO(3)$ to ensure gradient continuity. $\boldsymbol{\beta}$ denotes shape coefficients, while $d$ and $\mathbf{t}$ represent camera-space depth and 3D root translation, respectively. We utilize $L_1$ loss for its robustness to outliers.

\noindent\textbf{3D Supervision Loss.}
To ensure pose and shape fidelity, we supervise root-relative 3D joints $\mathbf{J}$ and mesh vertices $\mathbf{V}$:
\begin{align}
\mathcal{L}_{\text{3D}} &= \lambda_{J}\|(\hat{\mathbf{J}} - \hat{\mathbf{J}}_0) - (\mathbf{J} - \mathbf{J}_0)\|_1 + \lambda_{V}\|(\hat{\mathbf{V}} - \hat{\mathbf{J}}_0) - (\mathbf{V} - \mathbf{J}_0)\|_1,
\end{align}
where $\mathbf{J}_0$ is the pelvis root. This root-relative formulation decouples local body deformation from global localization, facilitating more stable learning of articulated poses.

\noindent\textbf{2D Reprojection Loss.}
To enforce alignment with image observations, we project 3D joints and vertices onto the 2D image plane using camera intrinsics $\mathbf{K}$:
\begin{equation}
\mathcal{L}_{\text{2D}} = \lambda_{j2d}\|\Pi(\hat{\mathbf{J}}) - \mathbf{j}_{2d}\|_1 + \lambda_{v2d}\|\Pi(\hat{\mathbf{V}}) - \mathbf{v}_{2d}\|_1,
\end{equation}
where $\Pi(\mathbf{X}) = [f_x \frac{X}{Z} + c_x, f_y \frac{Y}{Z} + c_y]^\top$. To handle occlusions, the loss is only computed for points within the image boundaries ($W, H$) via a validity mask:
\begin{equation}
\mathcal{L}_{j}^{\text{2D}} = \mathbf{1}[0 < j_{2d,x}^{(j)} < W] \cdot \mathbf{1}[0 < j_{2d,y}^{(j)} < H] \cdot \|\Pi(\hat{\mathbf{J}}^{(j)}) - \mathbf{j}_{2d}^{(j)}\|_1.
\end{equation}

\noindent\textbf{Training Schedule.}
We adopt a curriculum strategy where $\mathcal{L}_{\text{2D}}$ is enabled after 10 epochs. This allows the model to first establish a plausible 3D structure via $\mathcal{L}_{\text{param}}$ and $\mathcal{L}_{\text{3D}}$, mitigating depth ambiguities that often arise when 2D projection error dominates early training.

\subsection{Implementation Details}
\label{sec:suppl_implementation}

\noindent\textbf{Network Architecture.}
Our framework integrates two pre-trained backbones: (1) \textbf{DA3-Giant} (DINOv2-Giant) for depth and camera estimation, yielding $37 \times 37 \times 3072$ features; (2) \textbf{Multi-HMR-L} (DINOv2-Large) for mesh recovery, providing $64 \times 64 \times 1024$ features. Features are fused via a two-layer MLP ($[4096 \to 1024 \to 1024]$) with ReLU activations. The association module employs $P=36$ learnable queries ($D=1024$) and $H=8$ attention heads, followed by a 2-layer Transformer decoder.

\noindent\textbf{Two-Stage Training.}
\textit{Stage 1:} Pre-training on BEDLAM synthetic sequences (4 consecutive frames as pseudo-multi-view) for 50 epochs. We use AdamW with an effective batch size of 32. The learning rate follows a linear warmup ($10^{-8}$ to $10^{-4}$) followed by cosine decay.
\textit{Stage 2:} Fine-tuning on real multi-view EgoHumans data (2-4 synchronized views) for 50 epochs. We introduce cross-view reprojection losses ($\lambda_{\text{crossview\_j3d}}=10.0, \lambda_{\text{crossview\_j2d}}=1.0$) and increase weight decay to 0.01. Throughout both stages, backbones remain frozen while task-specific heads and fusion modules are optimized. All experiments are conducted on 4$\times$ NVIDIA H200 GPUs with distributed training.

\noindent\textbf{Loss Weights.}
For detection losses, we use $\lambda_{\text{bce}}=0.01$ and $\lambda_{\text{offset}}=1.0$. For SMPL parameter supervision, we set $\lambda_{\text{rotmat}}=0.1$, $\lambda_{\text{shape}}=1.0$, $\lambda_{\text{dist}}=1.0$, and $\lambda_{\text{transl}}=1.0$. The 3D supervision losses use $\lambda_{\text{j3d}}=100.0$ and $\lambda_{\text{v3d}}=100.0$, while the 2D reprojection losses use $\lambda_{\text{j2d}}=1.0$ and $\lambda_{\text{v2d}}=1.0$. For association losses, we set $\lambda_{\text{assign}}=1.0$ and $\lambda_{\text{contra}}=1.0$. In Stage 2, we additionally apply cross-view reprojection losses with $\lambda_{\text{crossview\_j3d}}=10.0$ and $\lambda_{\text{crossview\_j2d}}=1.0$.

\noindent\textbf{Cross-View Identity Association Algorithm.}
We provide the complete algorithmic description of our Cross-View Identity Association module. \cref{alg:association} describes the forward pass, which proceeds in six stages: (1) intra-view self-attention enhances each detection's representation by considering other people in the same view; (2) feature projection maps tokens to a normalized embedding space for contrastive learning; (3) cross-attention between learnable person queries and all detection tokens produces a soft assignment matrix; (4) weighted aggregation pools multi-view information per person; (5) greedy assignment maps detections to person identities; and (6) the Human Head receives aggregated features via both context injection (enhancing key-value features) and query injection (auxiliary attention token).

\begin{algorithm}[t]
\caption{Cross-View Identity Association (Forward Pass)}
\label{alg:association}
\begin{algorithmic}[1]
\REQUIRE Multi-view detection tokens $\{\mathbf{z}_n^v\}$ for $v \in \{1,...,S\}$, $n \in \{1,...,N_v\}$; Learnable person queries $\mathbf{Q} \in \mathbb{R}^{P \times D}$; Reference view features $\mathbf{F}^{\text{ref}} \in \mathbb{R}^{H \times W \times D}$
\ENSURE SMPL parameters $\{\boldsymbol{\theta}_p, \boldsymbol{\beta}_p, \mathbf{d}_p\}$ for each person $p$
\STATE \textcolor{gray}{\textit{// Stage 1: Intra-view self-attention enhancement}}
\FOR{each view $v \in \{1, ..., S\}$}
    \STATE $\tilde{\mathbf{Z}}^v \gets \text{LayerNorm}(\mathbf{Z}^v + \text{SelfAttn}(\mathbf{Z}^v, \mathbf{Z}^v, \mathbf{Z}^v))$
\ENDFOR
\STATE \textcolor{gray}{\textit{// Stage 2: Feature projection for contrastive learning}}
\STATE $\mathbf{H} \gets \text{LayerNorm}(\text{Proj}_{\text{contrast}}(\tilde{\mathbf{Z}}))$ \hfill $\triangleright$ $\mathbf{H} \in \mathbb{R}^{N \times D'}$, L2-normalized
\STATE \textcolor{gray}{\textit{// Stage 3: Cross-view soft assignment via cross-attention}}
\STATE $\mathbf{K} \gets [\tilde{\mathbf{z}}_1, ..., \tilde{\mathbf{z}}_N]^\top$ \hfill $\triangleright$ Concatenate all views
\STATE $\mathbf{A} \gets \frac{1}{H}\sum_{h=1}^{H}\text{softmax}\left(\frac{\mathbf{Q}\mathbf{W}_h^Q (\mathbf{K}\mathbf{W}_h^K)^\top}{\sqrt{D/H}}\right)$ \hfill $\triangleright$ $\mathbf{A} \in \mathbb{R}^{P \times N}$
\STATE \textcolor{gray}{\textit{// Stage 4: Multi-view feature aggregation}}
\FOR{each person query $p \in \{1, ..., P\}$}
    \STATE $\mathbf{f}_p^{\text{agg}} \gets \text{LayerNorm}(\mathbf{A}_{p,:} \cdot \mathbf{K})$ \hfill $\triangleright$ Weighted sum
\ENDFOR
\STATE \textcolor{gray}{\textit{// Stage 5: Person-to-detection assignment}}
\FOR{each detection $n$}
    \STATE $\text{person}(n) \gets \arg\max_{p} \mathbf{A}_{p,n}$
\ENDFOR
\STATE \textcolor{gray}{\textit{// Stage 6: Multi-view context injection into Human Head}}
\FOR{each person $p$ with assigned detections}
    \STATE $\tilde{\mathbf{C}} \gets \mathbf{C}^{\text{ref}} + \sigma(\alpha) \cdot \text{Broadcast}(\text{Proj}_{\text{kv}}(\mathbf{f}_p^{\text{agg}}))$ \hfill $\triangleright$ Context injection
    \STATE $\mathbf{q}_{\text{mv}} \gets \text{Proj}_{\text{q}}(\mathbf{f}_p^{\text{agg}})$ \hfill $\triangleright$ Query injection
    \STATE $\mathbf{q}_{\text{det}} \gets [\mathbf{z}_{\text{ref}}; \bar{\boldsymbol{\theta}}; \bar{\boldsymbol{\beta}}; \bar{\mathbf{c}}]$
    \STATE $\mathbf{q}_{\text{final}} \gets \text{CrossAttn}([\mathbf{q}_{\text{det}}; \mathbf{q}_{\text{mv}}], \tilde{\mathbf{C}}, \tilde{\mathbf{C}})$
    \STATE $\boldsymbol{\theta}_p, \boldsymbol{\beta}_p, \mathbf{d}_p \gets \text{MLP}_{\text{pose/shape/cam}}(\mathbf{q}_{\text{final}})$
\ENDFOR
\STATE \textbf{return} $\{\boldsymbol{\theta}_p, \boldsymbol{\beta}_p, \mathbf{d}_p, \mathbf{A}, \mathbf{H}\}$
\end{algorithmic}
\end{algorithm}

\noindent\textbf{Dynamic Query Selection.}
Although we initialize $P=36$ learnable person queries to accommodate scenes with many individuals, the actual number of queries used during inference adapts to scene complexity. Specifically, we compute $P_{\text{active}} = \min(\max_{v} N_v, P)$, where $N_v$ denotes the number of detected humans in view $v$. Only the first $P_{\text{active}}$ queries participate in cross-attention, reducing computational overhead for sparse scenes. Furthermore, after soft assignment, queries that receive no detection assignments (i.e., $\sum_n \mathbf{A}_{p,n} < \epsilon$ for all detections) are filtered out, ensuring that only queries corresponding to actual persons proceed to the Human Head for SMPL regression. This dynamic mechanism enables efficient handling of scenes ranging from single-person to crowded multi-person scenarios without architectural changes.

\noindent\textbf{Dynamic Query Selection.}
Although we initialize $P=36$ learnable person queries to accommodate scenes with many individuals, the actual number of queries used during inference adapts to scene complexity. Specifically, we compute $P_{\text{active}} = \min(\max_{v} N_v, P)$, where $N_v$ denotes the number of detected humans in view $v$. Only the first $P_{\text{active}}$ queries participate in cross-attention, reducing computational overhead for sparse scenes. Furthermore, after soft assignment, queries that receive no detection assignments (i.e., $\sum_n \mathbf{A}_{p,n} < \epsilon$ for all detections) are filtered out, ensuring that only queries corresponding to actual persons proceed to the Human Head for SMPL regression. This dynamic mechanism enables efficient handling of scenes ranging from single-person to crowded multi-person scenarios without architectural changes.

\noindent\textbf{Cross-View Association Loss.}
\cref{alg:loss} describes the loss computation, which combines three complementary objectives: (1) \textbf{soft assignment loss} directly supervises attention weights using Hungarian matching to align predicted and GT person identities, followed by cross-entropy; (2) \textbf{InfoNCE contrastive loss} pulls together features of the same person across different views while pushing apart different individuals; and (3) \textbf{cross-view reprojection loss} enforces geometric consistency by transforming predictions from the reference view to other views and comparing with GT in both 3D (root-relative) and 2D (projected) spaces.

\begin{algorithm}[t]
\caption{Cross-View Association Loss Computation}
\label{alg:loss}
\begin{algorithmic}[1]
\REQUIRE Soft assignment $\mathbf{A} \in \mathbb{R}^{P \times N}$; Projected features $\mathbf{H} \in \mathbb{R}^{N \times D'}$; Predicted joints $\{\mathbf{J}_p^{\text{ref}}\}$; GT correspondences $\{(p^*, \{(v, n)\})\}$; Camera extrinsics $\{\mathbf{C2W}^v, \mathbf{W2C}^v\}$
\ENSURE Total association loss $\mathcal{L}_{\text{assoc}}$
\STATE \textcolor{gray}{\textit{// \textbf{Loss 1: Soft Assignment Supervision}}}
\STATE $\mathbf{A}^* \gets \mathbf{0}^{P^* \times N}$ \hfill $\triangleright$ Construct GT assignment matrix
\FOR{each GT person $p^*$ with detections $\{(v_i, n_i)\}$}
    \FOR{each $(v_i, n_i)$}
        \STATE $\mathbf{A}^*_{p^*, \text{global}(v_i, n_i)} \gets 1$
    \ENDFOR
    \STATE $\mathbf{A}^*_{p^*,:} \gets \mathbf{A}^*_{p^*,:} / \|\mathbf{A}^*_{p^*,:}\|_1$ \hfill $\triangleright$ Row normalization
\ENDFOR
\STATE $\pi \gets \text{HungarianMatch}(\mathbf{A}, \mathbf{A}^*)$ \hfill $\triangleright$ Optimal assignment
\STATE $\mathcal{L}_{\text{assign}} \gets -\frac{1}{P^*} \sum_{p^*} \sum_{n} \mathbf{A}^*_{p^*,n} \log \mathbf{A}_{\pi(p^*),n}$
\STATE \textcolor{gray}{\textit{// \textbf{Loss 2: InfoNCE Contrastive Learning}}}
\STATE $\mathcal{P} \gets \emptyset$ \hfill $\triangleright$ Positive pairs: same person, different views
\FOR{each GT person $p^*$ with detections in $\geq 2$ views}
    \FOR{each pair $(n_i, n_j)$ where $v_i \neq v_j$}
        \STATE $\mathcal{P} \gets \mathcal{P} \cup \{(n_i, n_j)\}$
    \ENDFOR
\ENDFOR
\STATE $\mathcal{L}_{\text{contra}} \gets -\frac{1}{|\mathcal{P}|} \sum_{(i,j) \in \mathcal{P}} \log \frac{\exp(\mathbf{h}_i \cdot \mathbf{h}_j / \tau)}{\sum_{k \neq i} \exp(\mathbf{h}_i \cdot \mathbf{h}_k / \tau)}$
\STATE \textcolor{gray}{\textit{// \textbf{Loss 3: Cross-View Geometric Consistency}}}
\STATE $\mathcal{L}_{\text{reproj}} \gets 0$
\FOR{each person $p$ visible in reference view}
    \STATE $\mathbf{J}_{\text{world}} \gets \mathbf{C2W}^{\text{ref}} \cdot \mathbf{J}_p^{\text{ref}}$ \hfill $\triangleright$ Transform to world
    \FOR{each other view $v$ where person $p$ is visible}
        \STATE $\mathbf{J}_{\text{pred}}^v \gets \mathbf{W2C}^v \cdot \mathbf{J}_{\text{world}}$ \hfill $\triangleright$ Project to view $v$
        \STATE $\mathbf{J}_{\text{gt}}^v \gets$ GT joints in view $v$
        \STATE $\mathcal{L}_{\text{reproj}} \mathrel{+}= \|(\mathbf{J}_{\text{pred}}^v - \mathbf{J}_{\text{pred},0}^v) - (\mathbf{J}_{\text{gt}}^v - \mathbf{J}_{\text{gt},0}^v)\|_1$ \hfill $\triangleright$ Root-relative 3D
        \STATE $\mathcal{L}_{\text{reproj}} \mathrel{+}= \lambda_{\text{2d}} \|\Pi(\mathbf{J}_{\text{pred}}^v) - \mathbf{j}_{\text{gt}}^{v,\text{2d}}\|_1$ \hfill $\triangleright$ 2D reprojection
    \ENDFOR
\ENDFOR
\STATE \textcolor{gray}{\textit{// \textbf{Total Association Loss}}}
\STATE $\mathcal{L}_{\text{assoc}} \gets \lambda_1 \mathcal{L}_{\text{assign}} + \lambda_2 \mathcal{L}_{\text{contra}} + \lambda_3 \mathcal{L}_{\text{reproj}}$
\STATE \textbf{return} $\mathcal{L}_{\text{assoc}}$
\end{algorithmic}
\end{algorithm}

\subsection{Data Augmentation}
\label{sec:suppl_augmentation}

Maintaining geometric consistency across views and modalities (depth, camera, SMPL) is critical. Our augmentation pipeline is designed to preserve these relationships under transformation.

\noindent\textbf{Geometric Augmentations.}
We apply random rotation ($\pm 30^\circ$) and scaling ($[0.8, 1.2]$). For each rotation, we update camera extrinsics and transform SMPL global orientation/translation via forward kinematics to maintain camera-world alignment. Camera intrinsics are scaled proportionally to scaling factors. For cropping, we use the union bounding box of all humans with a 5\% random translation. Non-square crops are zero-padded to square before resizing to $896 \times 896$.

\noindent\textbf{Photometric Augmentations.}
We apply color jittering (brightness/contrast/saturation $\pm 0.5$, hue $\pm 0.1$) with 0.9 probability and grayscale conversion with 0.05 probability. In Stage 2, we use a \textit{co-jittering} strategy (ratio 0.3) to balance cross-view consistency with individual view diversity.

\noindent\textbf{Annotation Consistency.}
Post-augmentation, the principal point is re-centered to $(W/2, H/2)$ for a canonical camera model. While body-local pose and shape remain invariant, global orientation is transformed by the inverse rotation matrix around the Z-axis, and root translation is updated accordingly to ensure the projected mesh remains aligned with the augmented image.

\section{Experimental Details}
\label{sec:suppl_experiments}

\subsection{Evaluation Details}
\label{sec:suppl_eval}

We employ the following metrics to evaluate human pose estimation and camera localization. To ensure fair comparison, predicted results are aligned with ground truth via $SE(3)$ (rigid) or $Sim(3)$ (similarity) transformations as specified.

\subsubsection{Human Pose Metrics}

\noindent\textbf{W-MPJPE} (World Mean Per-Joint Position Error) measures the average Euclidean distance between predicted and ground-truth 3D joint positions in the world coordinate frame, after $SE(3)$ alignment of predicted camera trajectories to ground truth.

\noindent\textbf{PA-MPJPE} (Procrustes-Aligned MPJPE) computes the joint position error after per-person $Sim(3)$ alignment, which removes global translation, rotation, and scale. This metric evaluates local pose accuracy independent of camera estimation or body scale.

\noindent\textbf{GA-MPJPE} (Group-Aligned MPJPE) applies a single $Sim(3)$ alignment to all humans in a scene simultaneously, measuring their relative spatial configurations without considering absolute world positioning.

\subsubsection{Camera Localization Metrics}

\noindent\textbf{TE} (Translation Error) measures the mean Euclidean distance between predicted and ground-truth camera centers after $SE(3)$ alignment, evaluating metric accuracy of camera positions.

\noindent\textbf{s-TE} (Scale-aligned TE) is computed after $Sim(3)$ alignment, providing scale-invariant evaluation of camera position estimates.

\noindent\textbf{AE} (Angular Error) measures the average angular discrepancy (in degrees) between predicted and ground-truth pairwise camera orientations across the scene.

\noindent\textbf{CCA@$\tau$} (Camera Center Accuracy) measures the proportion of cameras localized within $\tau$\% of the overall scene scale after $SE(3)$ alignment. \textbf{s-CCA@$\tau$} denotes the scale-aligned version using $Sim(3)$.

\noindent\textbf{RRA@$\tau$} (Relative Rotation Accuracy) measures the percentage of camera pairs with relative rotation error within $\tau$ degrees.

\subsubsection{Evaluation Datasets}

\noindent\textbf{EgoHumans.}
We evaluate on the following sequences from EgoHumans. Table~\ref{tab:egohumans_sequences} summarizes the evaluation sequences and camera configurations.

\begin{table}[h]
\centering
\caption{EgoHumans evaluation sequences and camera configurations.}
\label{tab:egohumans_sequences}
\small
\begin{tabular}{l|c|c}
\toprule
\textbf{Activity} & \textbf{Sequences} & \textbf{Cameras} \\
\midrule
01\_tagging & 001--014 & 1, 4, 6, 8 \\
02\_lego & 001--006 & 2, 3, 4, 6 \\
03\_fencing & 001--014 & 4, 5, 10, 13 \\
04\_basketball & 001--014 & 1, 3, 4, 8 \\
05\_volleyball & 001--011 & 2, 4, 8, 11 \\
06\_badminton & 001--061 & 1, 2, 5, 7 \\
07\_tennis & 001--013 & 4, 9, 12, 20 \\
\bottomrule
\end{tabular}
\end{table}

\noindent\textbf{EgoExo4D.}
EgoExo4D scenes are typically captured using four to six RGB cameras and an egocentric device (Aria glasses). For our experiments and the baselines, we use only the RGB images from sequences with correct re-identification. Sequences containing ego-centric RGB views, such as helmet-mounted cameras, are excluded. We evaluate 182 videos from the validation set, sampling one random frame per video. The videos include ground-truth annotations for human poses, locations, and camera parameters.

Note that EgoExo4D uses COCO 17 keypoints, while our model outputs SMPL 24 joints. We evaluate on the common joints between them: shoulders, elbows, wrists, hips, knees, and ankles (12 joints total).

Table~\ref{tab:egoexo4d_sequences} lists all evaluation takes and frame indices.

\begin{table*}[t]
\centering
\caption{EgoExo4D evaluation takes and frame indices.}
\label{tab:egoexo4d_sequences}
\resizebox{\textwidth}{!}{%
\scriptsize
\begin{tabular}{l|p{5.8cm}|l|p{5.8cm}}
\toprule
\textbf{Dataset} & \textbf{Takes / Frame Indices} & \textbf{Dataset} & \textbf{Takes / Frame Indices} \\
\midrule
cmu\_soccer & 06\_3/1426, 12\_2/6807, 16\_2/6373
& unc\_basketball & 02-24-23\_01\_3/84, 02-24-23\_02\_10/466, 02-24-23\_02\_11/927, 03-30-23\_02\_10/45, 03-30-23\_02\_14/7, 03-30-23\_02\_15/40, 03-30-23\_02\_17/9, 03-30-23\_02\_18/20, 03-30-23\_02\_19/7, 03-30-23\_02\_4/107, 03-30-23\_02\_5/25, 03-30-23\_02\_7/1141 \\
\midrule
georgiatech\_bike & 06\_12/170, 06\_2/97, 06\_6/74, 06\_8/15, 07\_10/28, 07\_12/38, 07\_2/97, 07\_4/46, 07\_6/67, 07\_8/138, 14\_12/593, 14\_2/1214, 14\_6/575, 14\_8/97, 15\_2/1508, 15\_4/844, 15\_6/1103, 15\_8/3153, 16\_2/882, 16\_6/3031, 16\_8/1274
& uniandes\_basketball & 001\_23/768, 001\_24/1386, 001\_26/146, 001\_27/439, 003\_38/32, 004\_23/369, 004\_44/261, 004\_45/667 \\
\midrule
georgiatech\_covid & 02\_10/2227, 02\_12/6974, 02\_14/2926, 02\_2/67, 02\_4/67, 04\_10/999, 04\_12/6160, 04\_4/2996, 04\_6/4528, 06\_2/47, 06\_4/64, 18\_10/5524, 18\_12/3457, 18\_2/2413, 18\_4/3534, 18\_6/4389, 18\_8/458
& uniandes\_dance & 002\_11/201, 002\_2/439, 008\_29/276, 008\_30/166, 008\_31/31, 008\_32/11, 008\_33/1105, 008\_34/753, 008\_35/607, 008\_36/1045, 008\_37/913, 008\_38/706, 016\_10/841, 016\_11/279, 016\_12/932, 016\_13/453, 016\_14/951, 016\_30/577, 016\_31/1709, 016\_32/377, 016\_33/1158, 016\_36/1247, 016\_37/145, 016\_38/1416, 016\_39/399, 016\_3/1239, 016\_42/1406, 016\_43/1271, 016\_44/1268, 016\_45/838, 016\_6/1361, 016\_7/1040, 016\_8/1488, 017\_6/1592, 019\_17/1003, 019\_18/509, 019\_19/1537, 019\_20/1089, 019\_22/81, 019\_24/484, 019\_25/183, 019\_26/1814, 019\_27/283, 019\_28/1411, 019\_46/412, 019\_47/790, 019\_49/1617, 019\_51/481, 019\_52/875, 019\_54/766, 019\_55/679, 019\_56/561, 019\_57/1073, 019\_58/192, 024\_11/1619, 024\_12/104, 024\_13/1419, 024\_14/1180, 024\_15/378, 024\_16/1569, 024\_17/1317, 024\_45/844, 024\_47/732, 024\_48/261, 024\_49/325 \\
\midrule
iiith\_cooking & 59\_2/7795, 64\_2/298, 89\_6/1177, 90\_4/1383
& upenn\_Dance & 0706\_4\_2/2512, 0706\_4\_3/1277, 0706\_4\_4/1670, 0706\_4\_5/1904, 0713\_3\_2/164, 0713\_3\_3/586, 0713\_3\_4/0, 0713\_3\_5/243, 0713\_4\_2/125, 0713\_4\_3/1280, 0713\_4\_4/308, 0713\_4\_5/262, 0713\_5\_4/238, 0713\_5\_6/2534 \\
\midrule
iiith\_soccer & 015\_2/1610
& upenn\_Piano & 0721\_1\_2/140, 0721\_1\_3/648, 0722\_1\_2/83, 0727\_Partner\_Dance\_3\_1\_2/62 \\
\midrule
nus\_cpr & 12\_1/1338, 12\_2/76
& utokyo\_pcr & 2001\_29\_2/5799, 2001\_29\_4/3491, 2001\_29\_6/550, 2001\_30\_2/2121, 2001\_30\_4/1696, 2001\_32\_2/6641, 2001\_32\_4/6048 \\
\midrule
sfu\_basketball & 012\_10/774, 012\_12/399, 012\_2/945, 012\_3/1506, 012\_4/66, 012\_6/526, 012\_7/1581, 012\_8/329, 016\_2/247, 04\_8/209, 05\_22/1902, 05\_26/29, 09\_11/32, 09\_12/1114
& utokyo\_soccer & 8000\_43\_2/3262, 8000\_43\_4/3472, 8000\_43\_6/2781 \\
\midrule
sfu\_cooking & 028\_12/1049, 007\_7/77, 008\_3/4164, 008\_5/3559
& & \\
\midrule
sfu\_covid & 004\_2/2828, 004\_4/5360, 008\_16/1595
& & \\
\bottomrule
\end{tabular}%
}
\end{table*}

\subsection{Training Data Statistics}
\label{sec:suppl_data_stats}

\noindent\textbf{BEDLAM (Stage 1).}
To establish a robust foundation, we utilizestablish a robust foundation, we utilize 21 synthetic scenes from the BEDLAM dataset, encompassing diverse indoor and outdoor environments such such as offices, gyms, and sburbs. Each scene contains 250--1000 frames with 3--10 subjects per i-view constraints by sampling 4 consecutive temporal frames per training instance. Following the official protocol, we divide the dataset into training and validation sets, comprising 75\% and 20\% of the total 1M unique bounding boxes, respectively.

\noindent\textbf{EgoHumans (Stage 2).}
For Stage 2 training, we use sequences that are disjoint from the evaluation set. Specifically, we train on basketball (13 sequences), volleyball (11 sequences), badminton (61 sequences), and tennis (13 sequences) activities, while the evaluation is performed on tagging (14 sequences), lego assembly (6 sequences), and fencing (14 sequences) activities. This ensures zero overlap between training and evaluation data, allowing us to assess generalization to unseen activities and scenes. Each sequence is captured by 4--8 synchronized cameras, and we randomly sample 2--4 camera views per training sample.

\subsection{Additional Ablation Results}
\label{sec:suppl_ablation}

Table~\ref{tab:ablation_backbone} provides the complete numerical results for the backbone ablation study visualized in the main paper. We evaluate different combinations of DA3 backbone sizes (Large, Giant) and Multi-HMR configurations (input resolution and backbone size). The Giant DA3 backbone with 896$\times$896 Multi-HMR-Large (896\_L) achieves the best overall performance, balancing human pose accuracy and camera localization metrics.

\begin{table*}[t]
    \centering
    \caption{\textbf{Ablation study on backbone variants} on EgoHumans. \textbf{Best} and \underline{second best} results are highlighted.}
    \vspace{-0.2cm}
    \resizebox{1.00\textwidth}{!}{%
    \begin{tabular}{cc|ccc|ccccc}
        \toprule
        \multirow{2}{*}{\textbf{DA3}} & \multirow{2}{*}{\textbf{Multi-HMR}} & \multicolumn{3}{c|}{\textbf{Human Metrics}} & \multicolumn{5}{c}{\textbf{Camera Metrics}} \\
        & & W-MPJPE$\downarrow$ & GA-MPJPE$\downarrow$ & PA-MPJPE$\downarrow$ & TE$\downarrow$ & s-TE$\downarrow$ & AE$\downarrow$ & RRA@10$\uparrow$ & s-CCA@10$\uparrow$ \\
        \midrule
        Large & 672\_B & 0.94 & \underline{0.19} & \underline{0.09} & 4.46 & 0.24 & 2.69 & \textbf{1.00} & 0.94 \\
        Large & 672\_L & 0.92 & \textbf{0.18} & \textbf{0.08} & 4.45 & 0.23 & 2.69 & \underline{0.99} & 0.95 \\
        Large & 896\_L & \textbf{0.83} & \underline{0.19} & \underline{0.09} & 4.55 & 0.22 & 2.68 & \underline{0.99} & 0.96 \\
        \midrule
        Giant & 672\_B & 1.11 & \underline{0.19} & \underline{0.09} & \textbf{4.34} & \underline{0.19} & \underline{1.50} & \textbf{1.00} & 0.96 \\
        Giant & 672\_L & 1.10 & 0.20 & 0.10 & \underline{4.36} & \underline{0.19} & \underline{1.50} & \textbf{1.00} & \underline{0.97} \\
        \midrule
        \rowcolor{gray!20}
        \textbf{Giant} & \textbf{896\_L} & \underline{0.88} & \textbf{0.18} & \textbf{0.08} & 4.54 & \textbf{0.16} & \textbf{1.47} & \textbf{1.00} & \textbf{0.98} \\
        \bottomrule
    \end{tabular}%
    }
    \label{tab:ablation_backbone}
    \vspace{-10pt}
\end{table*}

Table~\ref{tab:ablation_num_views} provides the complete numerical results for the number of views ablation study visualized in the main paper. We evaluate the impact of increasing the number of input views from 1 to 4. With a single view, only human pose metrics are available. As the number of views increases, both human pose accuracy and camera localization improve, with 4 views achieving the best overall performance.

\begin{table*}[t]
    \centering
    \caption{\textbf{Ablation study on the number of views} on EgoHumans. \textbf{Best} and \underline{second best} results are highlighted.}
    \vspace{-0.2cm}
    \resizebox{1.00\textwidth}{!}{%
    \begin{tabular}{c|ccc|ccccc}
        \toprule
        \multirow{2}{*}{\textbf{Num. Views}} & \multicolumn{3}{c|}{\textbf{Human Metrics}} & \multicolumn{5}{c}{\textbf{Camera Metrics}} \\
        & W-MPJPE$\downarrow$ & GA-MPJPE$\downarrow$ & PA-MPJPE$\downarrow$ & TE$\downarrow$ & s-TE$\downarrow$ & AE$\downarrow$ & RRA@10$\uparrow$ & s-CCA@10$\uparrow$ \\
        \midrule
        1 & 1.54 & \textbf{0.16} & 0.16 & - & - & - & - & - \\
        2 & 0.93 & 0.25 & \underline{0.09} & \underline{4.72} & \underline{0.17} & 2.02 & \textbf{1.00} & \textbf{1.00} \\
        3 & \underline{0.90} & 0.23 & \textbf{0.08} & 4.79 & \underline{0.17} & \underline{1.63} & \textbf{1.00} & 0.87 \\
        \midrule
        \rowcolor{gray!20}
        \textbf{4} & \textbf{0.88} & \underline{0.18} & \textbf{0.08} & \textbf{4.54} & \textbf{0.16} & \textbf{1.47} & \textbf{1.00} & \underline{0.98} \\
        \bottomrule
    \end{tabular}%
    }
    \label{tab:ablation_num_views}
    \vspace{-10pt}
\end{table*}

\subsection{Additional Qualitative Results}
\label{sec:suppl_qualitative}

\begin{figure*}[t]
    \centering
    \includegraphics[width=0.95\textwidth]{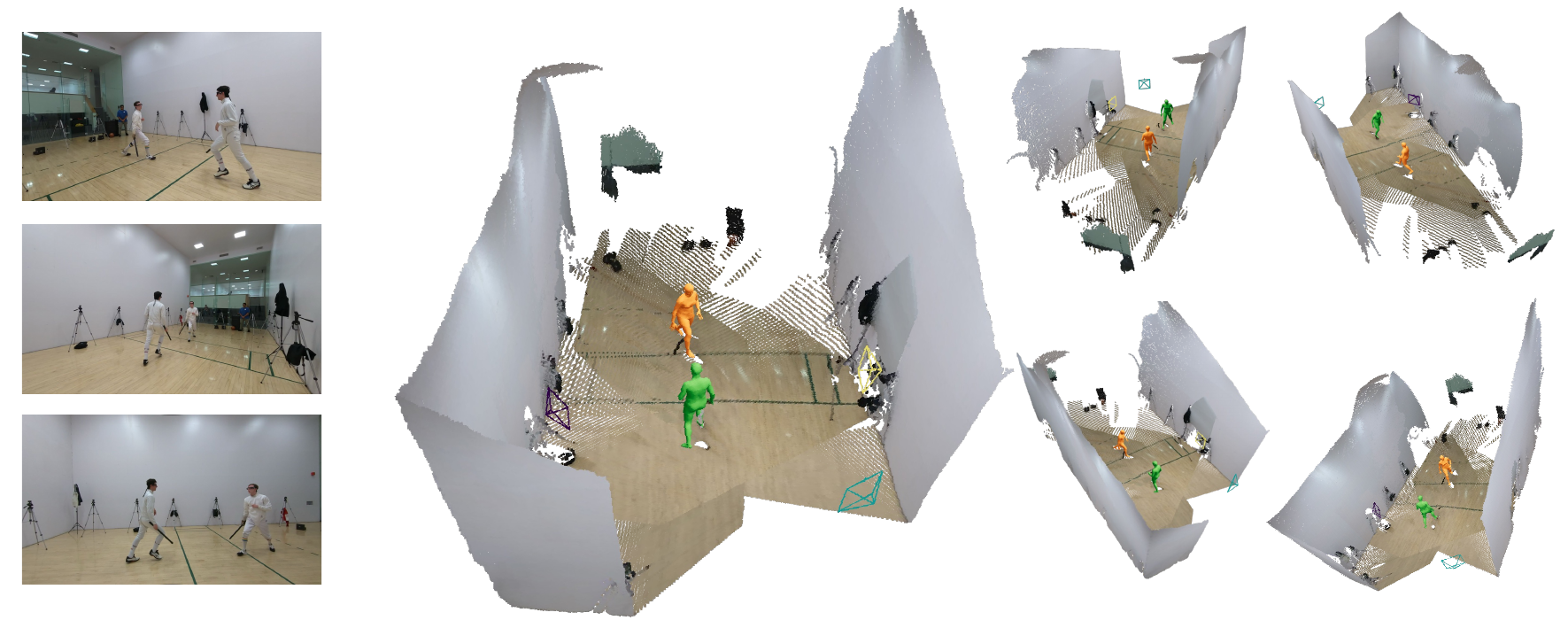}
    \vspace{-0.2cm}
    \caption{\textbf{Additional qualitative results.} Our method accurately reconstructs the 3D poses of two individuals (shown in orange and green) and recovers the camera poses from five different viewpoints. The scene point cloud and camera frustums are visualized together with the SMPL meshes, demonstrating consistent multi-view reconstruction.}
    \label{fig:case_supl_badminton}
\end{figure*}

\begin{figure*}[t]
    \centering
    \includegraphics[width=0.95\textwidth]{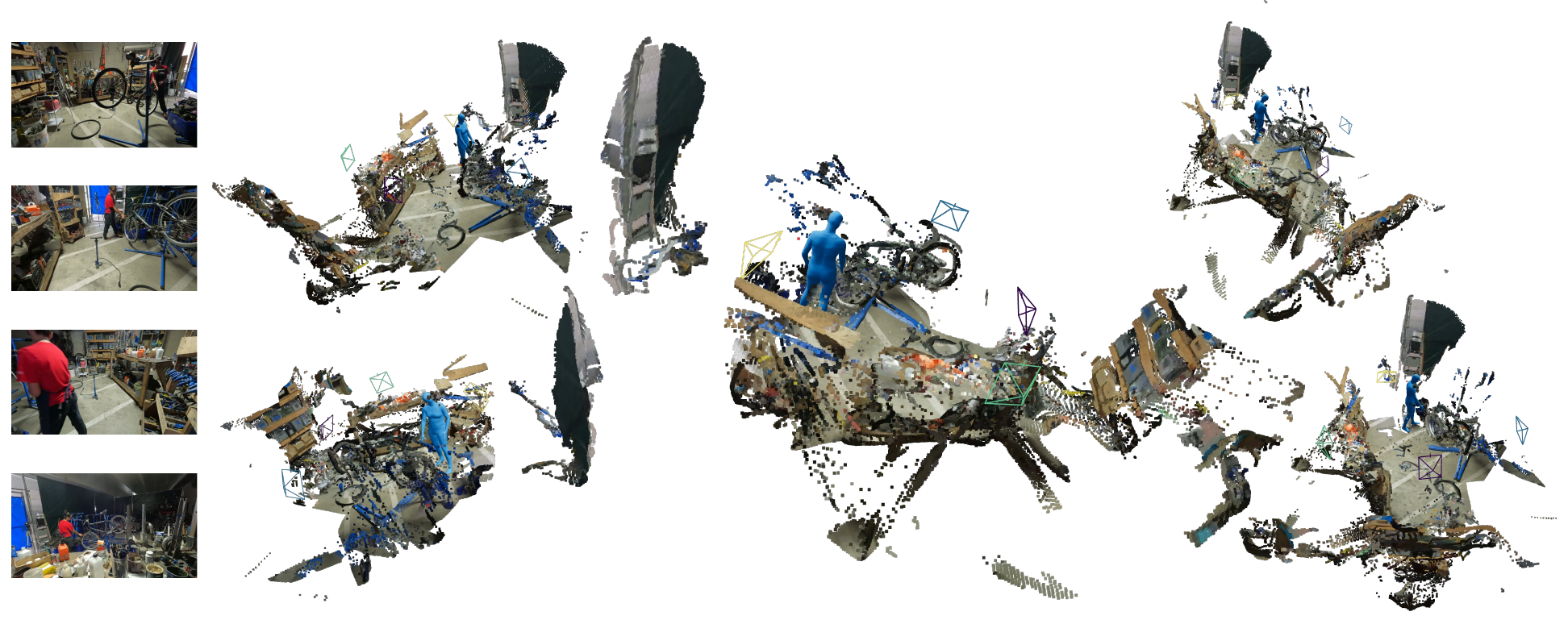}
    \vspace{-0.2cm}
    \caption{\textbf{Additional qualitative results.} Our method successfully handles complex real-world scenarios with cluttered backgrounds, diverse objects, and varying camera viewpoints. The reconstructed human (shown in blue) is accurately localized across 5 different camera views, demonstrating robustness to challenging scene conditions.}
    \label{fig:case_supl_egoexo}
\end{figure*}

We provide additional qualitative results to demonstrate the effectiveness of our method across diverse scenarios. \cref{fig:case_supl_badminton} shows multi-person reconstruction with accurate pose estimation and camera localization. \cref{fig:case_supl_egoexo} demonstrates robustness in cluttered environments with varying viewpoints.

\subsection{Failure Cases}
\label{sec:suppl_failure}

\begin{figure*}[t]
    \centering
    \includegraphics[width=0.95\textwidth]{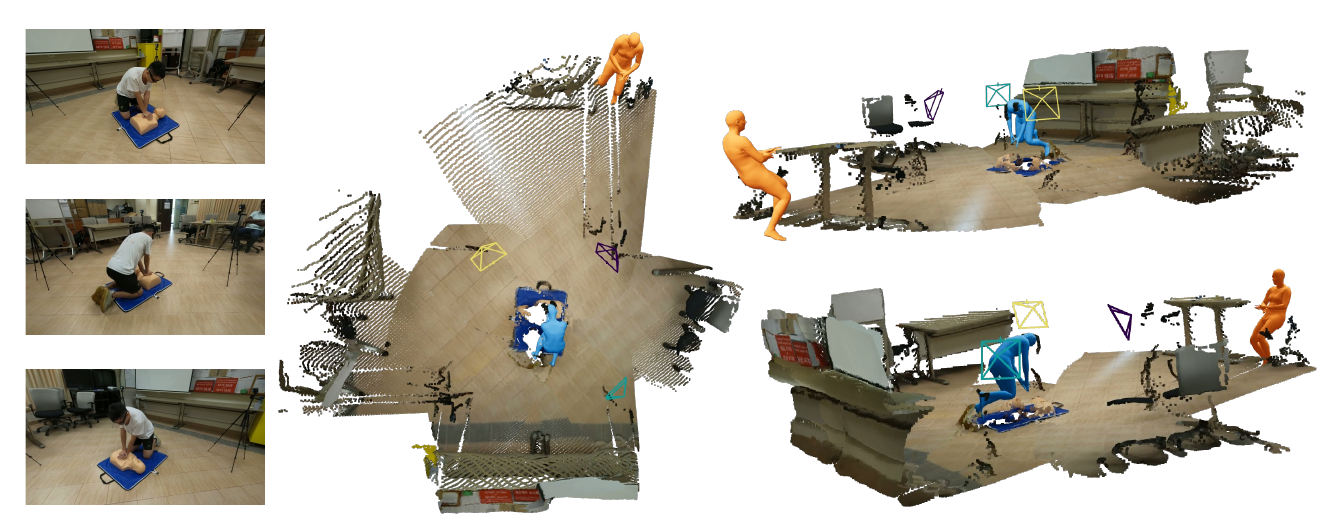}
    \vspace{-0.2cm}
    \caption{\textbf{Failure case.} The reconstructed human meshes appear floating above the ground plane, with feet not properly aligned with the floor surface.}
    \label{fig:case_failure}
\end{figure*}

We also present failure cases of our method. As shown in \cref{fig:case_failure}, the reconstructed humans appear floating above the ground.

\end{document}